\documentclass[preprint,12pt]{elsarticle}

\usepackage{amsmath,amssymb,amsfonts}
\usepackage{algorithmic}
\usepackage{graphicx}
\usepackage{multirow}
\usepackage{textcomp}
\usepackage{colortbl}
\usepackage{color}
\usepackage{xcolor}
\usepackage{setspace}
\journal{Pattern Recognition}

\begin{document}

\begin{frontmatter}

%% Title, authors and addresses

%% use the tnoteref command within \title for footnotes;
%% use the tnotetext command for theassociated footnote;
%% use the fnref command within \author or \address for footnotes;
%% use the fntext command for theassociated footnote;
%% use the corref command within \author for corresponding author footnotes;
%% use the cortext command for theassociated footnote;
%% use the ead command for the email address,
%% and the form \ead[url] for the home page:
%% \title{Title\tnoteref{label1}}
%% \tnotetext[label1]{}
%% \author{Name\corref{cor1}\fnref{label2}}
%% \ead{email address}
%% \ead[url]{home page}
%% \fntext[label2]{}
%% \cortext[cor1]{}
%% \affiliation{organization={},
%%             addressline={},
%%             city={},
%%             postcode={},
%%             state={},
%%             country={}}
%% \fntext[label3]{}

\title{Understanding and Improving CNNs with Complex Structure Tensor: A Biometrics Study}

%% use optional labels to link authors explicitly to addresses:
%% \author[label1,label2]{}
%% \affiliation[label1]{organization={},
%%             addressline={},
%%             city={},
%%             postcode={},
%%             state={},
%%             country={}}
%%
%% \affiliation[label2]{organization={},
%%             addressline={},
%%             city={},
%%             postcode={},
%%             state={},
%%             country={}}

\author{Kevin Hernandez-Diaz\corref{cor1}}
\author{Josef Bigun, Fernando Alonso-Fernandez}
\address{ISDD Lab, Halmstad University, Kristian IV:s väg 3, \\ Halmstad, 30250, Halland, Sweden}

% \affiliation[inst1]{organization={Halmstad University, ISDD Lab},%Department and Organization
%             addressline={Kristian IV:s väg 3}, 
%             city={Halmstad},
%             postcode={30250}, 
%             state={Halland},
%             country={Sweden}}

\cortext[cor1]{Corresponding author email: kevin.hernandez-diaz@hh.se}

\begin{abstract}
%% Text of abstract

Our study provides evidence that CNNs struggle to effectively extract orientation features. We show that the use of Complex Structure Tensor, which contains compact orientation features with certainties, as input to CNNs consistently improves identification accuracy compared to using grayscale inputs alone. Experiments also demonstrated that our inputs, which were provided by mini complex conv-nets, combined with reduced CNN sizes, outperformed full-fledged, prevailing CNN architectures. This suggests that the upfront use of orientation features in CNNs, a strategy seen in mammalian vision, not only mitigates their limitations but also enhances their explainability and relevance to thin-clients. Experiments were done on publicly available data sets comprising periocular images for biometric identification and verification (Close and Open World) using 6 State of the Art CNN architectures. We reduced SOA Equal Error Rate (EER) on the PolyU dataset by 5-26$\%$ depending on data and scenario.

\end{abstract}

%%Graphical abstract
% \begin{graphicalabstract}
% \includegraphics{grabs}
% \end{graphicalabstract}

%%Research highlights
% \begin{highlights}
% \item Research highlight 1
% \item Research highlight 2
% \end{highlights}


\end{frontmatter}

%% \linenumbers

%% main text
\section{Introduction}
\label{intro}

Computer Vision (CV) defines texture as a repetitive pattern \cite{tuceryan1993texture}. Extracting texture features amounts to finding a vector of parameters from subsets of the pattern to uniquely characterize it. Ideally, it is then necessary for the 
vector to be invariant to translation inside the same texture and different in another \cite{zhang2002brief}. 
The squared magnitude of Fourier Transform, FT, a.k.a. the power spectrum, of an image is invariant to image translations.  Consequently, the power spectrum of local images has been a main source of texture perception studies \cite{julesztexton}, and texture feature vectors \cite{HOG}\cite{tuceryan1993texture}\cite{LBP}. Even Convolutional Neural Networks (CNN) are biased towards texture information \cite{geirhos2018imagenettrained}, even showing that some first-layer ImageNet trained CNNs' filters have similar appearance as members of a Gabor filter-bank \cite{krizhevsky2012imagenet}. Since Hubel and Wiesel’s Nobel Prize-winning work, \cite{hubel59}, we know that mammalian vision is enabled by cells, each tuned to a unique orientation of translating bars in its visual field, and are upfront in the brain’s data flow, at the visual cortex. There exist cell layers that are invariant to translation, with tune-in angles uniformly and densely sampling $[0-\pi]$, like Gabor functions \cite{gabor} magnitude responses \cite{kulikowski81}. 
%Some cells are invariant to translation, whereas tune-in angles are uniform, Gabor function \cite{gabor} magnitude responses approximate this processing \cite{kulikowski81}. 

Texture plays a crucial role in many image-based biometric recognition systems such as the case of fingerprints \cite{jain2001fingerprint} or iris \cite{daugman2009iris}. In 2009, the authors of \cite{park2009periocular} showed the potential of the periocular region for biometric recognition. This region encompasses the iris, eyebrows, eyelids, commissures, skin texture, and the general eye's shape that can be used for recognition. This information-rich area has shown not only to be quite stable but also to provide high performance for soft-biometrics recognition \cite{alonso2016survey}. The periocular area provides flexibility regarding acquisition and occlusion as a middle ground between face and iris recognition. Like its biometric siblings, it has been the target of several texture extraction algorithms for recognition purposes with good performance \cite{periocularlbp}\cite{hernandez2023one}.

In this paper, we propose a set of complex mini conv-nets specialized in extracting texture presence and orientation with texture pattern complexity perfectly representable up to the n-th order moments of the power spectrum \cite{bigun06vd} in a region to enhance the performance of CNNs. The main contributions of this paper are as follows:

%In this paper, we propose a combination of a texture-extraction algorithm based on up to the Second Order Complex Derivative of Gaussians \cite{[Bigun04]} to detect the presence of linear symmetry patterns in a region to enhance the performance of CNNs. The main contributions of this paper are as follows:
\begin{itemize}
     \item We show that CNN networks can benefit from compact orientation features as texture features at the input by improving their a) explainability and focus, b) convergence, c) network size, and d) performance.
     \item We suggest complex Structure Tensors (CST), which is a further generalization of the ordinary Structure Tensor (ST), as a mini conv-net, relying on complex scalar products and complex non-linearities to provide compact inputs to CNNs. The features provided are dense equals of the local power spectrum, including one provided by a full Gabor filter bank, which approximates cell layers of mammalian vision well.
     \item Our evidence from biometric recognition confirms more explicitly previous indications that SOA CNN networks alone cannot produce orientation features significant for decisions on training data with reasonable size  ($ \approx$ ten thousand images). 
    \item Results indicate that the findings above generalize across different CNN architectures and spectra. 
 \end{itemize}

\section{Background and Related Work}\label{background}

This paper uses a method based on the Structure Tensor Theory (ST) \cite{bigun87london} as a compact texture orientation descriptor combined with CNNs. The descriptor, explained in detail in Section \ref{LST}, employs Complex Derivative of Gaussians to extract gradient information and detect the presence and orientation of line patterns. Besides biometrics, STs have been studied and used in numerous applications, including image enhancement, \cite{weickert1999coherence}, and medical image analysis \cite{szczepankiewicz2016link}, but also as a fundamental tool in computer vision, to detect and track key-point with \cite{lindeberg2013scale} or without scale-space \cite{mikolajczyk}, to generate dense optical flow, \cite{bruhn2005lucas}, as well as orientation maps \cite{si2017dense}.
Symmetry descriptors have been used previously in biometric recognition with success, given their resilience to viewpoint or scale changes \cite{symmetryfingerprint}, but their possible contribution has not yet been studied in the context of Deep Learning. Regarding periocular recognition, the works \cite{symmetryfingerprint} used the presence and orientation of several symmetry patterns at different scales around selected image keypoints for both periocular and forensic fingerprint recognition, achieving performance comparable to SOA.

Traditional methods like ST \cite{bigun87london}, LBP \cite{LBP}, HOG \cite{HOG}, and Gabor \cite{gabor} try to quantify the presence of line patterns in images, specifically their orientations. However, lately, their use has been superseded by CNNs. Even other studies on CNNs show the importance of quantifying orientation. While filtering profiles similar to Gabor filters are observed in the initial layers of CNNs \cite{gabornet}\cite{krizhevsky2012imagenet}, such filtering i) lacks systematic uniform orientation tuning and ii) the (non-linear) magnitude processing for translation invariance. This gap may arise from the assumption that all layers in CNNs must be determined by training data, whereas mammalian vision quantifies orientation uniformly up-front and maintains it throughout adult life.

To exploit potential synergies, researchers have tried to create robust models by combining traditional methods and CNNs in different ways. The authors in \cite{LBPCNNfeatures} combined LBP and RGB features with CNNs for face anti-spoofing, increasing performance while reducing the network size. In \cite{periocularlbp}, the authors proposed a double stream CNN that takes RGB ocular images and Orthogonal Combination-Local Binary Coded Pattern (OCLBCP) texture descriptor as inputs, improving recognition performance. In \cite{LBPNet}, the authors created a new CNN layer called Local Binary Convolution (LCB) as an alternative to convolutional layers. The network, called Local Binary Convolutional Neural Network (LBCNN), performed the same as regular CNNs while reducing network size and learnable parameters between 9 and 169 times.

In \cite{HOGCNNtracking}, the authors used HOG features and CNN to create an efficient pedestrian tracking and reidentification model across multiple camera views. The authors of \cite{hogcnnperiocular} used handcrafted HOG features, pre-trained CNNs features, and gender information extracted from a shallow network for periocular recognition. The work \cite{hogcnnexpression} used a HOG descriptor over spatial and temporal CNNs shallow features for face expression recognition in video.

The work \cite{gabornet} introduced GaborNet, a modified CNN in which the first layer filters are set to fit the Gabor Function by learning Gabor filter parameters instead of filter values directly. They showed that this modified CNN can outperform classic CNN while reducing the number of learnable parameters. 
The authors of \cite{gaborensemble} apply a bank of Gabor filters to face images to obtain responses. Each response image was used to train its own CNN network to achieve face classification. In \cite{gaboremotions}, the authors used the output of two consecutive Gabor filters to train a CNN to recognize between seven possible facial emotions. They showed that Gabor outputs can help CNNs achieve better emotion classification and faster convergence. The work \cite{gaborcapsule} showed that using the response of Gabor filters as input of CNNs can improve recognition of soft-biometrics recognition on face images in  CNNs and capsule networks.

%The authors of \cite{gaborensemble} proposed a Gabor CNN ensemble method for face recognition. Their method applies a bank of Gabor filters to extract the Gabor face representation, which is used to train a bank of CNNs, each trained for a specific Gabor response. In \cite{gaboremotions}, the authors used the output of two consecutive Gabor filters to train a CNN to recognize between seven possible facial emotions. They showed that Gabor outputs can help CNNs achieve better performance and faster convergence. The work \cite{gaborcapsule} showed that using the response of Gabor filters, which can enhance features such as wrinkles, as input of CNNs can improve the performance of soft-biometrics recognition on face images on both standard CNNs and capsule networks.

% text in GaborNet: In [13] was shown that deep CNN trained on real-life images tends to learn first convolutional layers contain mostly Gabor-like filters. Filters of the first layers of AlexNet is shown in Figure 2. This corroborates an idea of using Gabor filters in the first layer of CNN."

%The ensamble paper also provides very good sentences about CNNs and their limitations on the filters they can learn"

\section{Structure Tensor and power spectrum}
\label{LST}

\begin{figure}[h!]
\centering
\includegraphics[width=0.5\columnwidth]{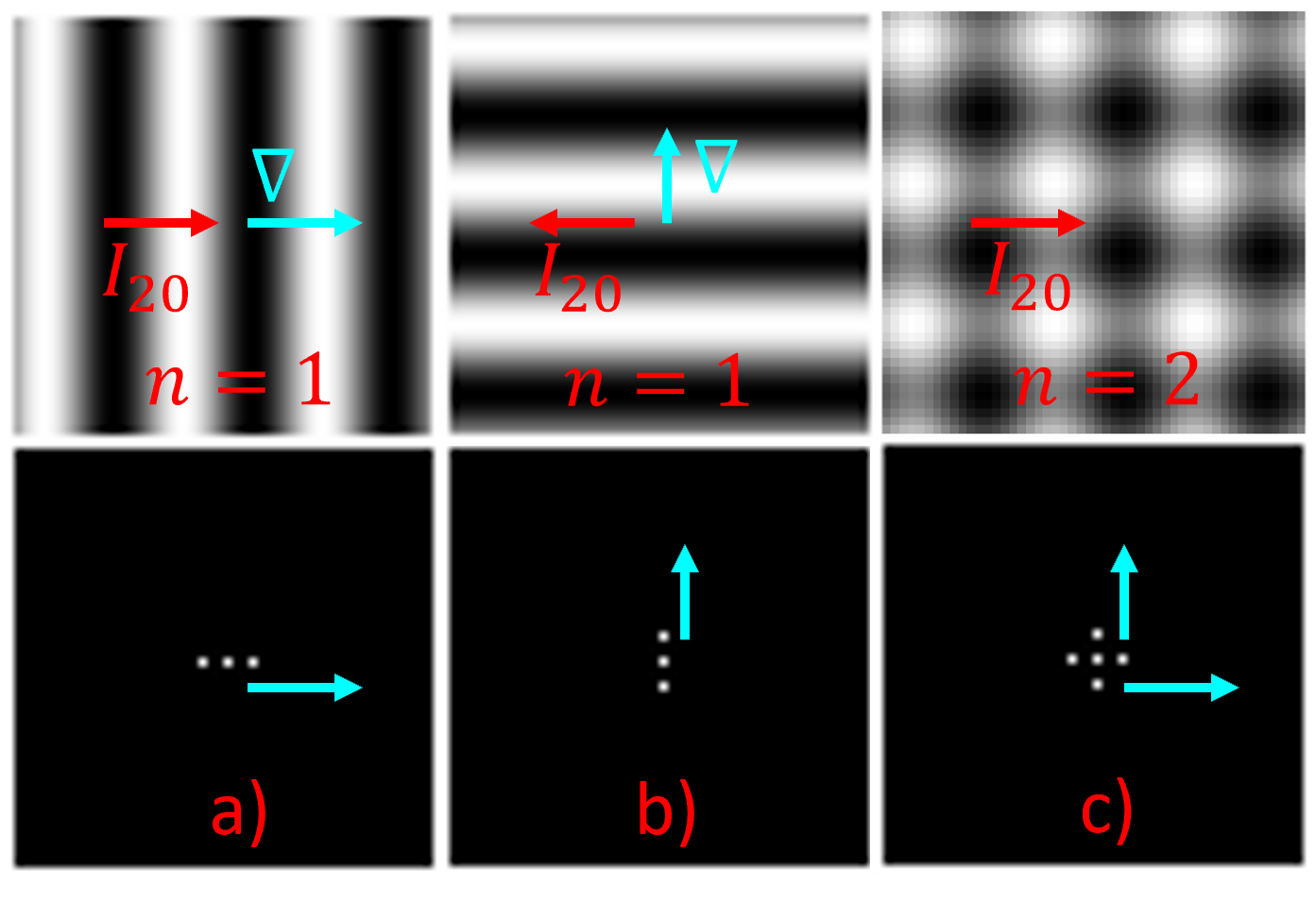}
\caption{a),b): Examples of  1-folded (linearly) symmetric textures of
planar waves with $ I_{20}$ and gradient vectors; c) dito but  2-folded symmetric, with $n=2$. The angle of $I_{20} $ (red arrows) is twice that of the gradient (blue arrows). FT magnitudes are below.  \label{fourier}}
\end{figure}

The Structure Tensor, ST \cite{bigun87london}, was originally introduced to provide a compact parametrization of the local power spectrum, which provides rich translation invariant features in texture. ST fits the total least square error line to the local spectrum and calculates the best and worst possible errors. Linearly symmetric patterns family in a local patch, those having iso-curves that are parallel, excite ST the most. In Figures \ref{fourier} a), b) two (local) patches exemplify the family, here defined as real parts of two distinct planar waves, $A\exp (i\omega_x x
+i\omega_yy) $.

$\mathbf{ST}$ is the $ 2\times2$ symmetric real semipositive definite matrix obtained by averaging the tensor products of image gradients $ \mathbf{ST} =\langle\nabla \! f \nabla^T\!\! f\rangle$ with a filter representing the neighborhood. Its three real elements encode the presence of an oriented texture in the neighborhood.
%, e.g. non-redundant elements of the $ \mathbf{ST}$.

Satisfying $\mathbf {ST}  u=\lambda u $, eigenvalues $0\le\lambda_2\le \lambda_1$ with corresponding (orthogonal) eigenvectors $ u_2$ and $ u_1 $ constitute another representation of ST, useful to explain the neighborhood in terms of measurable human observation, texture orientation.

The matrix $\mathbf {ST} $ can be decomposed as 
\begin{equation}
\mathbf{ST} =(\lambda_1-\lambda_2)u_1u_1^T+ \lambda_2
(u_1u_1^T+u_2u_2^T) \label{eq:spectral}
\end{equation}
where $ u_1u_1^T+u_2u_2^T=E$ is the unity matrix. Accordingly, $\mathbf {ST}$ can be uniquely decomposed into an oriented ``line'' component (encoded by $ u_1 u_1^T$) with weight-power of $\lambda_1-\lambda_2$ and a non-oriented, ``balanced'', component (encoded by identity) with the weight-power of $\lambda_2 $. 
%The tensor or its unique decomposition has been widely used in CV e.g. in tensor-voting \cite{medioni}, scale-space \cite{lindeberg98}, and diffusion \cite{weickert1998} processes.

%% The total available weight-power of all directions (=eigenvectors) is $\lambda_1+\lambda_2 $. When the neighborhood is linearly symmetric, then $ \lambda_2=0$.
%% Also, the representation of the neighborhood direction by $u_1u_1^T$, rather than $u_1 $, encodes continuously the fact that
%% %a neighborhood belonging to the family is invariant to rotation by $\pi$ radians since $ (-u_1)(-u_1)^T=u_1u_1^T$.
%% such patterns possess then 2-folded rotational symmetry, $ 2\pi/2$,in the neighborhood and in the spectrum. Equally important is that $\mathbf {ST} $ itself is invariant to translation so long as the neighborhood remains inside the texture (of the linearly symmetric pattern).

If and only if the neighborhood is linearly symmetric, i.e. it possesses texture orientation, its entire spectral power collapses to Dirac distributions located on a line through the (spectral) origin. In the solution, the error of the best fit is $ \lambda_2 $ (ideally zero for direction $u_1$), and the worst fit is $\lambda_1$ (for direction $u_2$). Accordingly, the difference $\lambda_1-\lambda_2 $
%is ``how much better'' the best error is w.r.t. the worst,
represents the confidence in the fit, whereas $\theta=\angle u_1 $ is the direction of the pattern. However, the confidence is energy dependent, and can vary with the neighborhood contrast.
%e.g., depending on the light producing the contrast (energy) of lines.

To exploit the ST theory more effectively, complex numbers can be used to represent the unique decomposition. 
%with complex numbers (instead of 2D vectors). Even gradients are thus encoded as complex numbers. 
The main advantage of this ``overhaul'' is the mathematical completability of STs by power series expansion enabled via the complex representation, allowing to express more elaborate oriented textures, those possessing n-folded symmetries.

Complex numbers reduce the unique decomposition  (\ref{eq:spectral}) to second order complex moments\footnote{
  $ I_{pq}\! \!\triangleq \!\!\int
  (\omega_x\!+\!i\omega_y)^p
  (\omega_x\!-\!i\omega_y)^q |F(\omega_x\!,\!\omega_y) |^2
  \!d\boldsymbol{\omega} $
  where $ F $ is FT of $ f $.
  }  of the neighborhood power spectrum, the complex $I_{20}  $, and real $ I_{11}  $,
\begin{equation}
\label{LST_eq}
     \mathbf{CST} \!= \!\begin{pmatrix}
I_{20},
  I_{11}
\end{pmatrix} \!=\!
\begin{pmatrix}
(\lambda_1 \! - \!\lambda_2)\exp({2i\angle u_1 }) ,\,
\lambda_1 \!+ \! \lambda_2
\end{pmatrix}
\end{equation}
Here $ i=\sqrt {-1} $,  and the $\mathbf {CST}  $ vector has 3 degrees of freedom. Components are readily connected to the unique decomposition of $ \mathbf {ST}$. The \textbf{CST} vector is thus the complex equivalent of \textbf{ST}.  Replacing the tensor mapping $ u_1u_1^T$ in 2D, the complex mapping $ 2\angle u_1 $ is similarly invariant to pattern rotation with $ \pi $ i.e. $2(\angle u_1+\pi)=2\angle u_1 $.

The sum $\lambda_1+\lambda_2=I_{11} $ is a tight upper bound for the confidence $0\le \lambda_1-\lambda_2 \le  \lambda_1+\lambda_2$ as $\lambda_2 $ approaches to zero. An exploitable fact is then, that confidence $\left | I_{20} \right |$ equals the upper bound $I_{11}$ if and only if the pattern possesses texture orientation (=linearly symmetric). 
%% It has been shown that,
%% %% \cite{bigun87london}, LST elements are   second order
%% %% complex moments of the local power spectrum.
%% Since the power spectrum is even, there can only be two uniquely defined, independent complex moments of order 2, $I_{20} $ and $ I_{11} $, that is $ I_{02} $ is non-informative due to $I_{02}=I_{20}^*   $.

% \Figure[ht!](topskip=0pt, botskip=0pt, midskip=0pt)[width=0.95\textwidth]{pipeline_josef_26.png}
% {Pipeline of the proposed method. \label{I20process}}

\begin{figure}
\centering
\includegraphics[width=0.95\columnwidth]{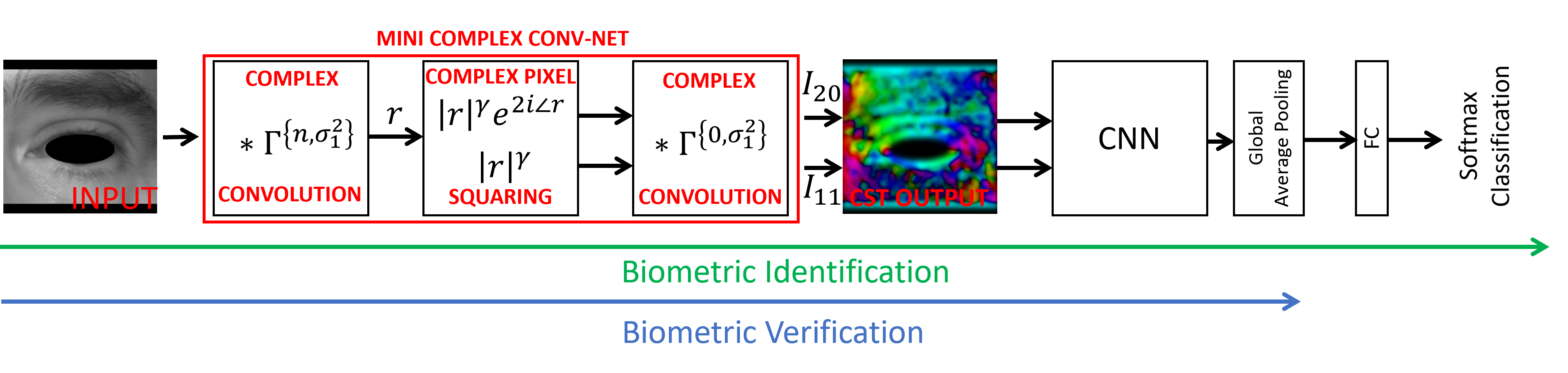}
\caption{Pipeline of the proposed method. \label{I20process}}
\end{figure}

The first benefit is that we can estimate these moments with just 3 separable filters, i.e., without a massive filtering process (with a  Gabor filter bank), producing the local power spectrum after taking magnitude responses. This is because the axis fitting process is directly available in $\mathbf{CST}  $.
As illustrated by Figure \ref{I20process}, the moments can be obtained directly via a mini-complex convolutional network, comprising convolution of the original image with complex filters given by
\begin{equation}
\label{eq:gammader}
\Gamma^{ \{ n, \sigma^2\} }(x,y)= (D_x+iD_y)^n G_{\sigma^2} (x,y)\}
\end{equation}
where $D_x$, $D_y$ are partial derivatives, and
\begin{equation}
G_{\sigma^2} (x,y)=\frac {1} {2\pi \sigma^2}
\exp(- \frac {x^2+y^2} {2\sigma^2} ),
\end{equation}
is the familiar Gaussian function with variance $\sigma^2 $.  In the case of texture orientation of linear symmetry type  ($ n=1 $), the first (linear) box becomes the gradient computation.

%% We will return shortly to (\ref{eq:gammader}) when the integer $ n>1 $
%% in the first box, i.e. when the complex gradient is  no-longer the
%% ordinary gradient.
%% %% a.k.a.  symmetry
%% %%  derivative of order $ n $,
%% In the second box, we always have $ n=0 $ producing the translation
%% invariance of complex gradients.
%% %% By
%% %% setting $ n=1 $ and $ \sigma=\sigma_1 $, the first level complex filter $  \Gamma^{\{1,
%% %%   \sigma_1^2 $ is obtained which is $ -(x+iy)*G_{\sigma_1^2}
%% %%   /\sigma_1^2$.  The second level filter is obtained similarly by $\Gamma^{\{0, \sigma_2^2 $
%% %%   and is
%% %%   an ordinary Gaussian.
The {\bf linear symmetry algorithm} generates elements of \textbf{CST}, via three
steps each, as follows.
\begin{enumerate}
  \item Convolve gray image $ f $  with the filter $
\Gamma^{\{n,\sigma^2\} } $, by setting $ n=1 $, and $ \sigma=\sigma_1 $.
The result is a complex valued image, $ r $.
  \item Square each pixel in $ r $ while emphasizing (gradient)
magnitudes with $ \gamma$ to yield two  outputs, complex $r_a=|r|^\gamma
\exp(2\angle r) $ having double gradient
angles, and real $ r_b=|r|^\gamma $

%%    Emphasize magnitudes by exponentiation, i.e.  $
%%    |r|^{\gamma/2}  $, and   square  complex
%%    pixelwise,
%%    resulting in $|r|^{\gamma}\exp(i2\angle r)   $ which  doubles
%%     angles of
%%     $ r$.
  \item[3a] Convolve the $r_a $ strand of Step 2 with the Gaussian
filter $ \Gamma^{\{n,\sigma^2\} } $  where  $ n=0 $, and $
\sigma=\sigma_2 $ to provide the complex $ I_{20}  $.
  \item[3b] Convolve the $r_b $ strand of Step 2 with the same Gaussian
filter to output the real  $ I_{11}  $.
%% \item  Convolve the {\em magnitude} of the result of Step 2, i.e. $
%%   |r|^\gamma$,   with the same Gaussian   as in Step 3, resulting in
%%  the real image with non-negative pixel values representing $ I_{11}  $.
\end{enumerate}

%% All convolutions can be implemented by using separable, iterative and/or pyramidal filtering,  using classical signal processing techniques for any $\sigma $. 
Textures with unique orientation have their 2D power spectra concentrated into a line, as illustrated in Figure \ref{fourier} a), b).
However, multiple orientations can occur in textures such as repeated squarish patterns, as exemplified by Figure \ref{fourier} c), which will then produce a cross-like pattern in the power spectrum. This makes  $I_{20}$ with $ n=1 $ insufficient, requiring more sophisticated models of orientation presence.

The texture model represented in the above algorithm can then be
completed in the mathematical sense by increasing $ n $ to  $ n=2 $ in
the first box, \cite{bigun06vd}. This changes only
the used filter in the box to  $  \Gamma^{ \{2,\sigma^2\} }$,
(\ref{eq:gammader}), and the same algorithm will instead compute
elements of the vector ($ I_{2n,0} $, $I_{n,n}  $), completing the
descriptive ability of lower "$ n$"s. Such increases yield even orders of complex
moments of the (local or Gabor) power spectrum, which is even, to
quantify its {\em $ n$-folded symmetry} concentration. Instead of 1
axis as in $ n=1$, each $(I_{2n,0},I_{n,n} ) $ pair fits $ n $ evenly
distributed axes through the origin. The orientation delivered in $I_{2n,0}$ is texture group orientation, e.g., how many radians the lines of a squarish
pattern deviate from a reference direction.

%% The fitting is achieved by remapping the polar angle coordinates of the FT domain in the $n\rightarrow 1 $ mapping which is inherent to $\Gamma^{n, \sigma^2}  $. The mapping equates polar angles of the spectrum which are separated by $ \pi/n $ radians, aligning them along the same axis through the
%% origin. The subsequent steps, squaring and convolution, fit a single axis to the remapped power spectrum as in the case of $ n=1 $. Consequently, in the sequel we use terms like ``$ (I_{20}, I_{11} )  $ {\em with }  $n=2 $''  to mean $ (I_{40}, I_{22} ) $ of the power spectrum, since the multiple axes fitting is a single axis fitting underthe $ n\rightarrow 1 $ mapping of the first step.

Using $ \mathbf {CST} $,  we present evidence that CNNs are not able to reach the valuable information embedded in texture orientations. Likewise, the evidence supports that if the orientation of power spectrum concentrations is extracted and presented to CNNs via complex moments, then they can make decisions with improved accuracy.

\section{Methodology}\label{Method}

% \begin{figure}[!h]
% \centering
% \includegraphics[width=0.7\columnwidth]{sigmas_final.png}
% \caption{Effect of parameters $\sigma_1$ and $\sigma_2$ on the CST representation when using as input the top left image shown in Figure \ref{databases_image}. The CST images are shown in HSV color format. Hue is modulated according to the $I_{20}$ angle to correspond to the CIE color standard, with red being $0$ degrees orientation. Saturation is modulated according to $I_{11}$, and Value is modulated by the magnitude of $I_{20}$. \label{sigmaspic}}
% \end{figure}

\begin{figure}[!h]
\centering
\includegraphics[width=0.99\columnwidth]{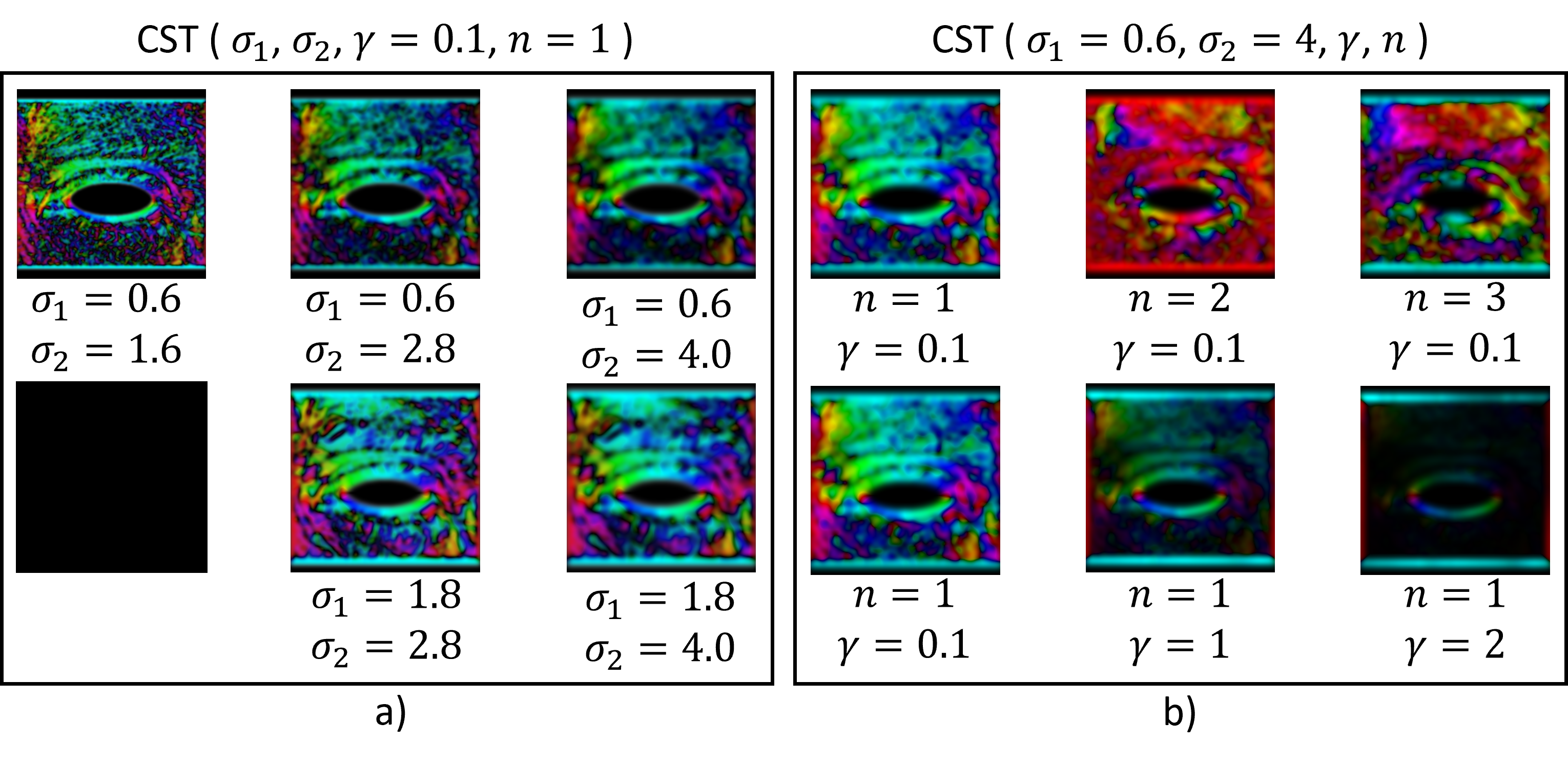}
\caption{Effect of hyperparameters in the CST output, shown in HSV-colors where Hue is modulated by $\angle I_{20}$ (i.e. $0^\circ$ is mapped to red), saturation by $I_{11}$, and Value by  $|I_{20}|$. When the latter two are high (vivid colors), certainty is high. \label{hyperparameters}}
\end{figure}

We follow the pipeline shown in Figure \ref{I20process} to introduce the CST into CNN architectures.
The  $ \sigma_1 $ of the first step determines the 
frequency band in which CST acts, whereas $ \sigma_2 $ in the third step determines neighborhood size. Figure \ref{hyperparameters} a) shows the effect of using different $\sigma$ values.
Preliminary results showed us that the highest frequencies in our
datasets contained the most discriminative information. The neighborhood size only
influenced methods where alignment was an issue. We settled therefore
for a value for $\sigma_1$ of $0.6$ to extract high-frequency
gradients and $\sigma_2$ of $4.0$ to
define minimum neighborhood size. Furthermore, we set $\gamma$ to $0.1$ to alleviate the difference in magnitudes between the regions with subtle and fast changes. Figure \ref{hyperparameters} b) shows the type of texture extracted according to different values of $n$ and $\gamma$ for the selected $\sigma_1$ and $\sigma_2$.

% \begin{figure}[!h]
% \centering
% \includegraphics[width=0.7\columnwidth]{./orderandgamma_jb2.png}
% \caption{Effect of symmetry order $n$ and $\gamma$ emphasis on CST outputs, shown in HSV-colors where Hue is modulated by $\angle I_{20}$ (i.e. $0^\circ$ is mapped to red), saturation by $I_{11}$, and Value by  $|I_{20}|$. \label{ordergammapic}}
% \end{figure}

% To corroborate  conclusions, 
%  we  carried out  biometric recognition 
%  experiments by publicly  available six
%  SOA CNN networks,  ResNet50 \cite{resnet}, DenseNet121
%  \cite{densenet}, VGG16 \cite{vgg}, Xception \cite{xception},
%  InceptionV3 \cite{inception}, MobileNetV2 \cite{mobilenet}.  

To corroborate conclusions, we carried out biometric recognition experiments by six widely used CNN architectures:  ResNet50 \cite{resnet}, DenseNet121 \cite{densenet}, Xception \cite{xception}, InceptionV3 \cite{inception}, MobileNetV2 \cite{mobilenet}.

% We used Tensorflow-Keras as our Deep Learning framework to download, initialize, train, and test the models. The CST process was transparent to the network and not affected by the training process. We use the standard models provided by the Keras library, only changing the last fully connected layer to fit the number of users in the database and the input size to fit the type of data used. We trained the networks using Stochastic Gradient Descent with a learning rate of $0.001$ and a momentum of $0.9$, except for VGG16, which did not use momentum and had a Clip-Value of $0.5$. The networks were trained for $100$ epochs with a batch size of $16$. No data augmentation was used during training. Model checkpoint was used to monitor the validation loss and recover the best-performing version after training. All training was conducted on a Windows 10 machine with an i7-8700, 32GB of RAM, and an Nvidia Titan V. 

We used Tensorflow-Keras as our Deep Learning framework to download, initialize, train, and test the models. The CST process was transparent to the network and not affected by the training process. We use the standard models provided by the Keras library, only changing the last fully connected layer to fit the number of users in the database and the input size to fit the type of data used. We trained the networks using Stochastic Gradient Descent with a learning rate of $0.001$ and a momentum of $0.9$. The networks were trained for $100$ epochs with a batch size of $16$. No data augmentation was used during training. Model checkpoint was used to monitor the validation loss and recover the best-performing version after training. All training was conducted on a Windows 10 machine with an i7-8700, 32GB of RAM, and an Nvidia Titan V.

\begin{table*}[ht!]
\centering
\resizebox{0.95\textwidth}{!}{%
\begin{tabular}{|c|c|c|c|c|c|c|}
\hline
\multirow{2}{*}{\textbf{Database}} &
  \multirow{2}{*}{\textbf{Protocol}} &
  \multirow{2}{*}{\textbf{\begin{tabular}[|c|]{@{}c@{}} Images \\ (Classes)\end{tabular}}} &
  \multirow{2}{*}{\textbf{\begin{tabular}[c|]{@{}c@{}}Train Parition\\Images (Classes)\end{tabular}}} &
  \multirow{2}{*}{\textbf{Gen./Imp.Pairs}} &
  \multirow{2}{*}{\textbf{\begin{tabular}[c|]{@{}c@{}}Test Partition\\Images (Classes)\end{tabular}}} &
  \multirow{2}{*}{\textbf{Gen./Imp.Pairs}} \\
  & & & & & &\\ \hline
\multirow{3}{*}{\textbf{PolyU}}      & \textbf{5-fold} & \multirow{3}{*}{6,270 (418)} & 5,016 (418)        & -                & 1,254 (418)    & -                \\ \cline{2-2} \cline{4-7} 
                                     & \textbf{CW}     &                              & 4,180 (418)        & 18,810/8,715,300 & 2,090 (418)    & 4,180/2,178,825  \\ \cline{2-2} \cline{4-7} 
                                     & \textbf{OW}     &                              & 3,135 (209)        & 21,945/4,890,600 & 3,135 (209)    & 2,1945/4,890,600 \\ \hline
\multirow{3}{*}{\textbf{Cross-Eyed}} & \textbf{5-fold} & \multirow{3}{*}{1920 (240)}  & *1,680/1,440 (240) & -                & *240/480 (240) & -                \\ \cline{2-2} \cline{4-7} 
                                     & \textbf{CW}     &                              & 1,200 (240)        & 2,400/717,000    & 720 (240)      & 720/258,120      \\ \cline{2-2} \cline{4-7} 
                                     & \textbf{OW}     &                              & 960 (120)          & 3,360/456,960    & 960 (120)      & 3,360/456,960    \\ \hline
\end{tabular}
}
\caption{Summary of Train/Test partitions per database, spectrum, and protocol for the verification experiments, Both databases contain the same number of images in each spectrum. The Close-World (CW) and Open-World (OW) protocols with PolyU and Cross-Eyed are defined following \cite{depresion}. $^*$The values indicate the number of images when the test fold is composed of 1 or 2 images per class due to the number of images per user not being divisible by 5.}\label{tabla-gen-imp}
\end{table*}

\section{Databases, Metrics, and Protocol}
\label{DBMetricsProt}

\subsection{Databases}
\label{datasets}

\begin{figure}[!h]
\centering
\includegraphics[width=0.6\columnwidth]{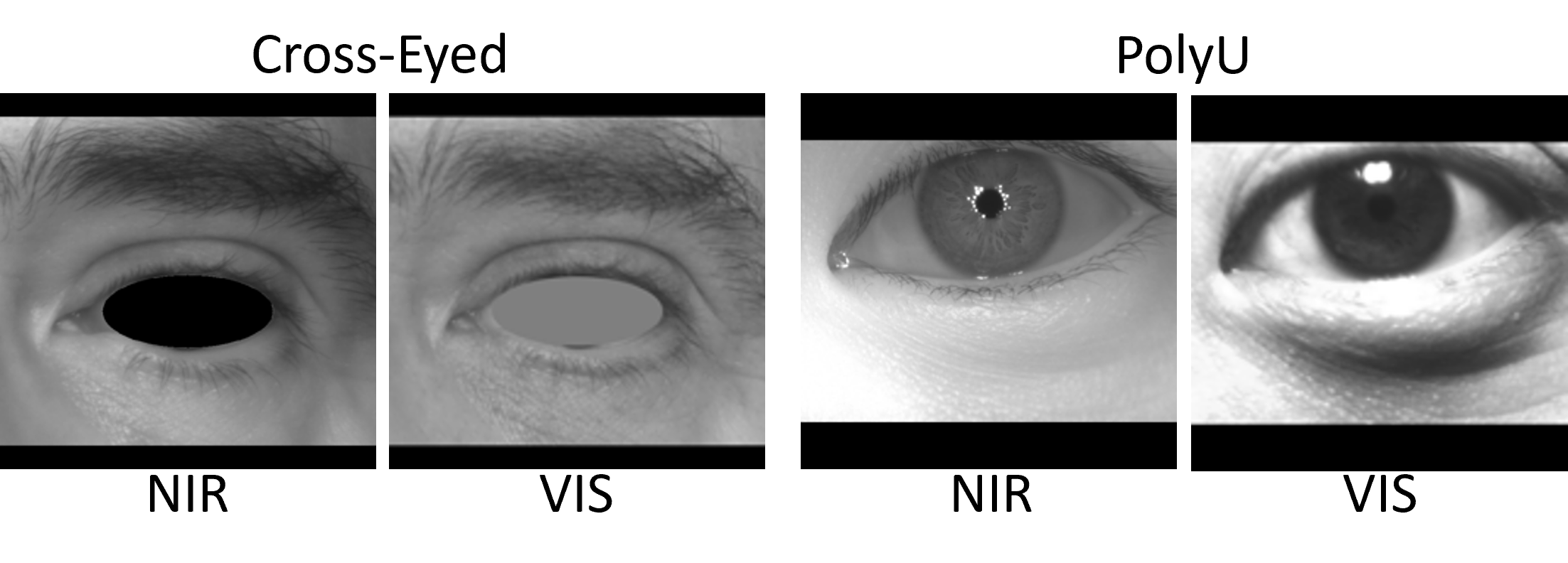}
\caption{Example images from the databases employed.  \label{databases_image}}
\end{figure}

We employed two commonly used periocular datasets in the experimentation: Cross-Eyed \cite{x-eyed2016}, and PolyU \cite{polyu}. Figure \ref{databases_image} shows some examples of each database.

The \textbf{Cross-Eyed} dataset is a cross-spectral periocular database that offers Near-Infrared (NIR) and Visible (VIS) images from subjects of different nationalities, ethnicities, and eye colors using a custom dual-spectrum image sensor under normal indoor illumination. The database contains $8$ images of each eye from $120$ subjects per spectrum, for a total of $3,840$ images. Images were normalized to have the same sclera radius, center, orientation, size, and dimensions.

% For each of the $120$ subjects, they captured $8$ images of both eyes in both spectra for a total of $3840$ images. They provide a version with the sclera masked for the periocular-only challenge to ensure no iris information was used. We used the iris masks to normalize the images to have the same sclera radius, center, and orientation. Finally, they were zero-padded and cropped, so all have the same size and dimensions.

The \textbf{PolyU} database is a bi-spectral iris database captured using simultaneous bi-spectral imaging by the Hong Kong Polytechnic University. It provides images in NIR and VIS, captured simultaneously, offering exact pixel correspondence between both spectral image versions. The database consists of $15$ instances for each spectrum of both eyes of $209$ different subjects for a total of $12,540$ iris images. Since the periocular region in this dataset is smaller, images are zero-padded to be square while maintaining the aspect ratio.

\subsection{Metrics and Protocol}
\label{metricsprotocol}

To evaluate the performance of the proposed system, we carried out both biometric identification and verification experiments. For the identification cases, we did 5-fold cross-validation and tested across all user samples using softmax and cross-entropy loss. We performed biometric verification to frame our methods across the SOA results for the used databases. We followed the same protocols as in \cite{depresion}: the Close-World (CW) protocol, in which the training and test partition contains images of all users without overlap, and the Open-World (OW) protocol, in which the users are split into training and testing along with all their images. The networks used for biometric verification were previously trained for identification using the same training and validation partitions. For testing, we extracted the features from the second to last layer and compared them using the cosine similarity.

\section{Results and Discussion}\label{Discuss}

\subsection{Effect of CST input variables}

\begin{table*}[h!]
\centering
\resizebox{0.75\textwidth}{!}{%
\begin{tabular}{c|c|c|c|c|c|c|c|c|c|c|} \cline{2-11}

      &\multicolumn{6}{|c|}{\textbf{Input}} &
  \multicolumn{2}{|c|}{\textbf{Cross-Eyed}} &
  \multicolumn{2}{|c|}{\textbf{PolyU}} \\ \cline{2-11}  
      &\multicolumn{1}{|c|}{\textbf{BW}} &
  \multicolumn{1}{|c|}{\textbf{$\left | I_{20} \right |$}} &
  \multicolumn{1}{|c|}{\textbf{$\angle I_{20}$}} &
  \multicolumn{1}{|c|}{\textbf{$\Re(I_{20})$}} &
  \multicolumn{1}{|c|}{\textbf{$\Im(I_{20})$}} &
  \textbf{$I_{11}$} &
  \multicolumn{1}{|c|}{\textbf{NIR}} &
  \textbf{VIS} &
  \multicolumn{1}{|c|}{\textbf{NIR}} &
  \textbf{VIS} \\ \hline  
   \multicolumn{1}{|c|}{\textbf{$1^{st}$}}   &\multicolumn{1}{|c|}{\textbf{X}} &
  \multicolumn{1}{|c|}{\textbf{}} &
  \multicolumn{1}{|c|}{\textbf{}} &
  \multicolumn{1}{|c|}{\textbf{}} &
  \multicolumn{1}{|c|}{\textbf{}} &
  \textbf{} &
  \multicolumn{1}{|c|}{97.8} &
  97.7 &
  \multicolumn{1}{|c|}{93.2} &
  94.5 \\ \hline  
  \multicolumn{1}{|c|}{\textbf{$2^{nd}$}}    &\multicolumn{1}{|c|}{\textbf{}} &
  \multicolumn{1}{|c|}{\textbf{X}} &
  \multicolumn{1}{|c|}{\textbf{}} &
  \multicolumn{1}{|c|}{\textbf{}} &
  \multicolumn{1}{|c|}{\textbf{}} &
  \textbf{} &
  \multicolumn{1}{|c|}{96.4} &
  97.1 &
  \multicolumn{1}{|c|}{93.7} &
  94.3 \\ \hline  
  \multicolumn{1}{|c|}{\textbf{$3^{rd}$}}    &\multicolumn{1}{|c|}{\textbf{}} &
  \multicolumn{1}{|c|}{\textbf{X}} &
  \multicolumn{1}{|c|}{\textbf{}} &
  \multicolumn{1}{|c|}{\textbf{}} &
  \multicolumn{1}{|c|}{\textbf{}} &
  \textbf{X} &
  \multicolumn{1}{|c|}{97.6} &
  97.6 &
  \multicolumn{1}{|c|}{94.7} &
  95.4 \\ \hline  
   \multicolumn{1}{|c|}{\textbf{$4^{th}$}}   &\multicolumn{1}{|c|}{\textbf{}} &
  \multicolumn{1}{|c|}{\textbf{X}} &
  \multicolumn{1}{|c|}{\textbf{X}} &
  \multicolumn{1}{|c|}{\textbf{}} &
  \multicolumn{1}{|c|}{\textbf{}} &
  \textbf{X} &
  \multicolumn{1}{|c|}{98.4} &
  98.4 &
  \multicolumn{1}{|c|}{95.6} &
  96.1 \\ \hline  
  \multicolumn{1}{|c|}{\textbf{$5^{th}$}}    &\multicolumn{1}{|c|}{\textbf{}} &
  \multicolumn{1}{|c|}{\textbf{}} &
  \multicolumn{1}{|c|}{\textbf{}} &
  \multicolumn{1}{|c|}{\textbf{X}} &
  \multicolumn{1}{|c|}{\textbf{X}} &
  \textbf{} &
  \multicolumn{1}{|c|}{98.2} &
  98.5 &
  \multicolumn{1}{|c|}{96.0} &
  96.0 \\ \hline  
  \multicolumn{1}{|c|}{\textbf{$6^{th}$}}    &\multicolumn{1}{|c|}{\textbf{}} &
  \multicolumn{1}{|c|}{\textbf{}} &
  \multicolumn{1}{|c|}{\textbf{}} &
  \multicolumn{1}{|c|}{\textbf{X}} &
  \multicolumn{1}{|c|}{\textbf{X}} &
  \textbf{X} &
  \multicolumn{1}{|c|}{98.5} &
  98.6 &
  \multicolumn{1}{|c|}{96.3} &
  \textbf{96.4} \\ \hline  
   \multicolumn{1}{|c|}{\textbf{$7^{th}$}}   &\multicolumn{1}{|c|}{\textbf{}} &
  \multicolumn{1}{|c|}{\textbf{X}} &
  \multicolumn{1}{|c|}{\textbf{}} &
  \multicolumn{1}{|c|}{\textbf{X}} &
  \multicolumn{1}{|c|}{\textbf{X}} &
  \textbf{X} &
  \multicolumn{1}{|c|}{98.4} &
  \textbf{98.7} &
  \multicolumn{1}{|c|}{\textbf{96.5}} &
  \textbf{96.4} \\ \hline  
   \multicolumn{1}{|c|}{\textbf{$8^{th}$}}   &\multicolumn{1}{|c|}{\textbf{X}} &
  \multicolumn{1}{|c|}{\textbf{}} &
  \multicolumn{1}{|c|}{\textbf{}} &
  \multicolumn{1}{|c|}{\textbf{X}} &
  \multicolumn{1}{|c|}{\textbf{X}} &
  \textbf{X} &
  \multicolumn{1}{|c|}{\textbf{98.6}} &
  98.2 &
  \multicolumn{1}{|c|}{\textbf{96.5}} &
  96.2 \\ \hline  
  \multicolumn{1}{|c|}{\textbf{$9^{th}$}}    &\multicolumn{1}{|c|}{\textbf{X}} &
  \multicolumn{1}{|c|}{\textbf{X}} &
  \multicolumn{1}{|c|}{\textbf{}} &
  \multicolumn{1}{|c|}{\textbf{X}} &
  \multicolumn{1}{|c|}{\textbf{X}} &
  \textbf{X} &
  \multicolumn{1}{|c|}{98.3} &
  98.3 &
  \multicolumn{1}{|c|}{\textbf{96.5}} &
  96.2 \\ \hline 
\end{tabular}
}
\caption{\label{tab:input_type} Average test accuracy from the 5-fold cross-validation identification experiments with respect to the variable type used at the input of a randomly initialized ResNet50.
}
\end{table*}

%\begin{figure}[!h]
%\centering
%\includegraphics[width=0.7\columnwidth]{inputs.png}
%\caption{Example of different CST representations computed from the bottom left image shown in Figure \ref{databases_image}. \label{inputs}}
%\end{figure}

\begin{figure}[!h]
\centering
\includegraphics[width=0.95\columnwidth]{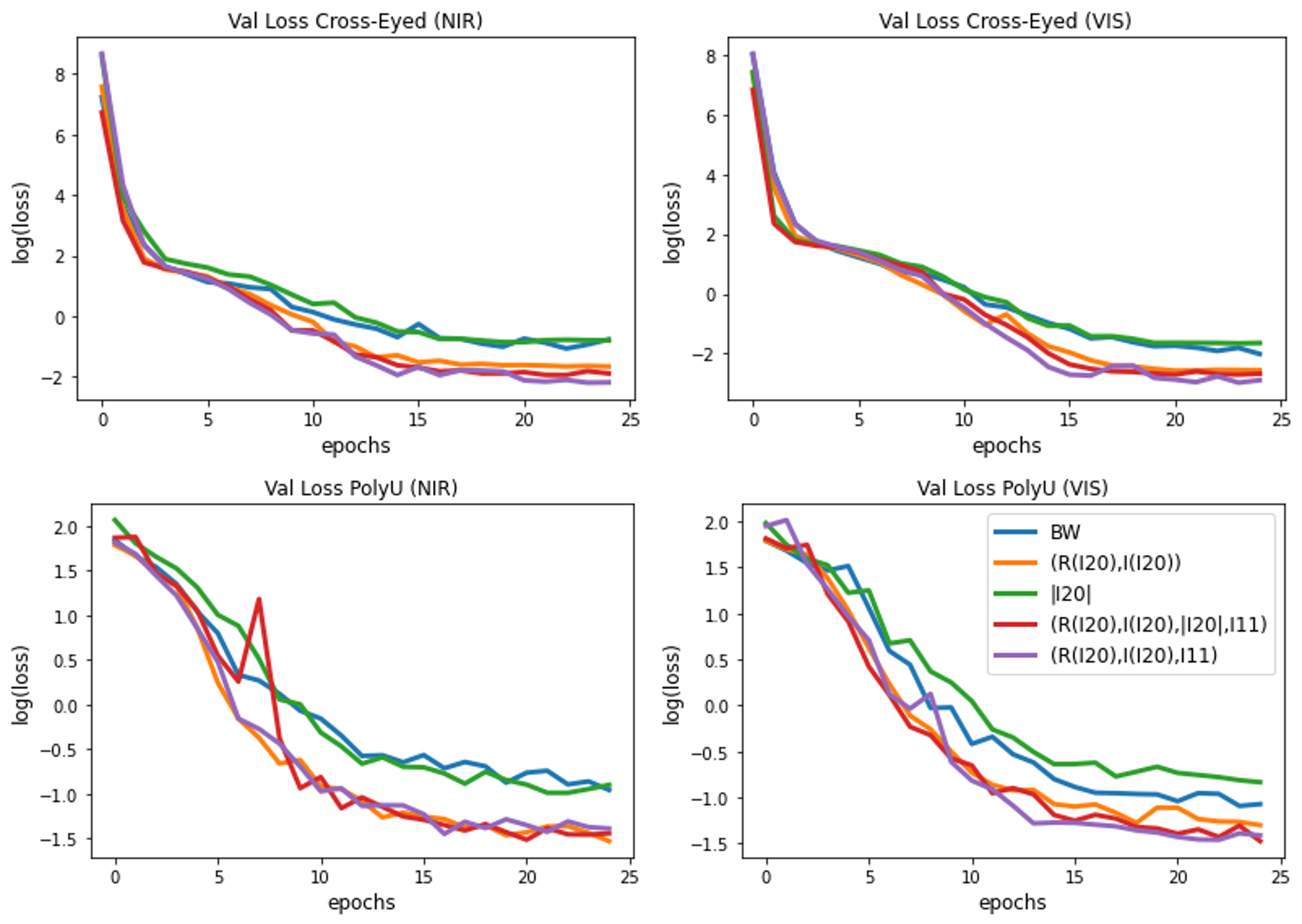}
\caption{Validation loss during training for a randomly initialized ResNet50 network using different input data. \label{converge_cross-eyed}}
\end{figure}

% \begin{figure}[!h]
% \centering
% \includegraphics[width=0.95\columnwidth]{converge_cross-eyed.png}
% \caption{Cross-Eyed database: validation loss and accuracy during training for a randomly initialized ResNet50 network using different input data. \label{converge_cross-eyed}}
% \end{figure}

% \begin{figure}[!h]
% \centering
% \includegraphics[width=0.95\columnwidth]{converge_cross-polyu.png}
% \caption{PolyU database: validation loss and accuracy during training for a randomly initialized ResNet50 network using different input data. \label{converge_polyu}}
% \end{figure}

The first step in our experimentation was to investigate how the representation and presence of $I_{20}$ and $I_{11}$ affected the training and performance of the network. Since $I_{20}$ is complex, we can use either the real $\Re(I_{20})$ and imaginary $\Im(I_{20}$ representation or the polar coordinates $\left | I_{20} \right |$  and $\angle(I_{20})$. The network input was modified by concatenating the selected representation through the channel axis. Table \ref{tab:input_type} presents the identification accuracy results for these experiments. Additionally, Figure \ref{converge_cross-eyed} shows the validation loss during the training process.

From Table \ref{tab:input_type}, we can see that the incremental improvement observed with the addition of each component in the polar coordinates indicates that magnitude, angle, and upper fitting boundary all contain beneficial information offering comparable or better performance than the original grayscale image. Furthermore, the angle information (fourth row) improves the accuracy further for all cases, outperforming the grayscale version for both datasets. This shows that the network can benefit from the orientation information upfront, even if the angle is discontinuous at $0/360$ degrees in the polar representation. 
Providing the real and imaginary parts of $I_{20}$ alone, which also implicitly encodes magnitude and angle, gives similar or better results than the polar counterparts, plus $I_{11}$. The addition of $I_{11}$ to this new case (sixth row) gives extra improvement, highlighting the valuable contribution of all elements in either representation. 
We also have studied the combination of the original grayscale image with the best variants of the CST representation (8th and 9th row). While also providing the grayscale image can help improve the accuracy, this may not hold true for all the options (sixth vs. ninth row). 
The best overall results are obtained with the real and imaginary parts, $\left | I_{20} \right |$, and $I_{11}$ (seventh row). This case provides not only direct information on the presence of the linear pattern in the area, but the orientation information is also provided by the real and imaginary values. Another consideration of these results is that our observations hold for the two databases and the two spectra, indicating the generalization capabilities of the proposed method.

We then analyze the benefits of the proposed methods during network training (Figure \ref{converge_cross-eyed}). We do so for selected cases in Table \ref{tab:input_type}. We can observe that just providing the real and imaginary parts already allows for faster convergence and lower loss than when using the grayscale version. However, as with the results in Table \ref{tab:input_type}, providing just the magnitude $\left | I_{20} \right |$ is insufficient to observe tangible benefits. Again, the best case overall corresponds with the addition of $I_{11}$ and $\left | I_{20} \right |$ to $I_{20}$, which can be observed regardless of the database and spectrum employed.

% For the remainder of this paper, we retain the configuration of $\Re(I_{20})$,$\Im(I_{20})$, $I_{11}$ as input to the network (sixth row of Table \ref{tab:input_type}). This input provides the best balance between the accuracy of the system and input size, having a standard number of channels for CNNs. Furthermore, the difference with the best-performing option (row seven) is only marginal and only for the PolyU database, which can also be affected by the stochastic nature of the network initialization and training.

\subsection{Effect of CST derivative order}

\begin{table}[h!]
\centering
\resizebox{0.65\textwidth}{!}{%
\begin{tabular}{|cccc|cc|cc|}
\hline
\multicolumn{4}{|c|}{\textbf{Input}} &
  \multicolumn{2}{c|}{\textbf{Cross-Eyed}} &
  \multicolumn{2}{c|}{\textbf{PolyU}} \\ \hline
\multicolumn{1}{|c|}{\textbf{BW}} &
  \multicolumn{1}{c|}{\textbf{$\Gamma ^{\left \{ 1,\sigma_{1}^2  \right \}}$}} &
  \multicolumn{1}{c|}{\textbf{$\Gamma ^{\left \{ 2,\sigma_{1}^2  \right \}}$}} &
  \textbf{$\Gamma ^{\left \{ 3,\sigma_{1}^2  \right \}}$} &
  \multicolumn{1}{c|}{\textbf{NIR}} &
  \textbf{VIS} &
  \multicolumn{1}{c|}{\textbf{NIR}} &
  \textbf{VIS} \\ \hline
\multicolumn{1}{|c|}{\textbf{X}} &
  \multicolumn{1}{c|}{\textbf{}} &
  \multicolumn{1}{c|}{\textbf{}} &
  \textbf{} &
  \multicolumn{1}{c|}{97.8} &
  97.7 &
  \multicolumn{1}{c|}{93.3} &
  94.5 \\ \hline
\multicolumn{1}{|c|}{\textbf{}} &
  \multicolumn{1}{c|}{\textbf{X}} &
  \multicolumn{1}{c|}{\textbf{}} &
  \textbf{} &
  \multicolumn{1}{c|}{98.6} &
  98.3 &
  \multicolumn{1}{c|}{95.9} &
  96.4 \\ \hline
\multicolumn{1}{|c|}{\textbf{}} &
  \multicolumn{1}{c|}{\textbf{}} &
  \multicolumn{1}{c|}{\textbf{X}} &
  \textbf{} &
  \multicolumn{1}{c|}{98.0} &
  97.7 &
  \multicolumn{1}{c|}{95.8} &
  95.5 \\ \hline
\multicolumn{1}{|c|}{\textbf{}} &
  \multicolumn{1}{c|}{\textbf{}} &
  \multicolumn{1}{c|}{\textbf{}} &
  \textbf{X} &
  \multicolumn{1}{c|}{97.3} &
  96.6 &
  \multicolumn{1}{c|}{93.7} &
  94.0 \\ \hline
\multicolumn{1}{|c|}{\textbf{}} &
  \multicolumn{1}{c|}{\textbf{X}} &
  \multicolumn{1}{c|}{\textbf{X}} &
  \textbf{} &
  \multicolumn{1}{c|}{98.2} &
  98.4 &
  \multicolumn{1}{c|}{96.0} &
  95.7 \\ \hline
\multicolumn{1}{|c|}{\textbf{}} &
  \multicolumn{1}{c|}{\textbf{X}} &
  \multicolumn{1}{c|}{\textbf{X}} &
  \textbf{X} &
  \multicolumn{1}{c|}{98.0} &
  98.5 &
  \multicolumn{1}{c|}{95.9} &
  96.0 \\ \hline
\end{tabular}
}
\caption{\label{tab:best_order} Average test accuracy for each combination of the Complex Derivative of Gaussian order $n$. The complex data representation used at the input is $\Re(I_{20})$,$\Im(I_{20})$, $I_{11}$.
}
\end{table}

Another relevant parameter related to the network’s input is how the symmetry order ($n$ in Equation \ref{eq:gammader}) affects the network. The higher the order, the more intricate texture orientation the model will extract (Figure \ref{hyperparameters}). However, the presence of highly complex patterns is dataset and application dependent. Nonetheless, higher-order symmetries still have some response, making an interesting case for the selection of the input data. In this paper, we experiment with values of $n$ from $1$ to $3$. For simplicity, for each value of $n$, we use the input combination consisting of $\Re(I_{20})$, $\Im(I_{20})$, and $I_{11}$. Table \ref{tab:best_order} shows the results of these experiments. We can see that when only one is used, the first-order derivative of Gaussian ($n=1$) performs the best, although the second-order still matches or improves the grayscale version. In general, the performance decreases as we increase $n$. This is an indication that classes could be separated already by low-order orientation quantification. Even though combining different orders improved the accuracy in some database/method combinations, the effect was not consistently observable. %Using only the first order provides the best accuracy, or the second-best accuracy by just one or two decimal points. This means that even though higher orders provide extra information, they may also add noise to the system. However, different behavior could happen for another type of data or task.

\subsection{Effect of network compression}
\label{compression}

% \Figure[h!](topskip=0pt, botskip=0pt, midskip=0pt)[width=0.9\textwidth]{parameters2.png}
% {Average test accuracy for different network compression levels. The number of parameters depends on the width of the network, controlled by coefficient $\alpha$, and the depth of the network, as explained in Section \ref{compression}. The complex data representation used at the input is $\Re(I_{20})$,$\Im(I_{20})$, $I_{11}$. 
% %parameter effect (point is 256 full size network and alpha=0.2.
% \label{parametereffect}}

\begin{figure}[!h]
\centering
\includegraphics[width=0.95\columnwidth]{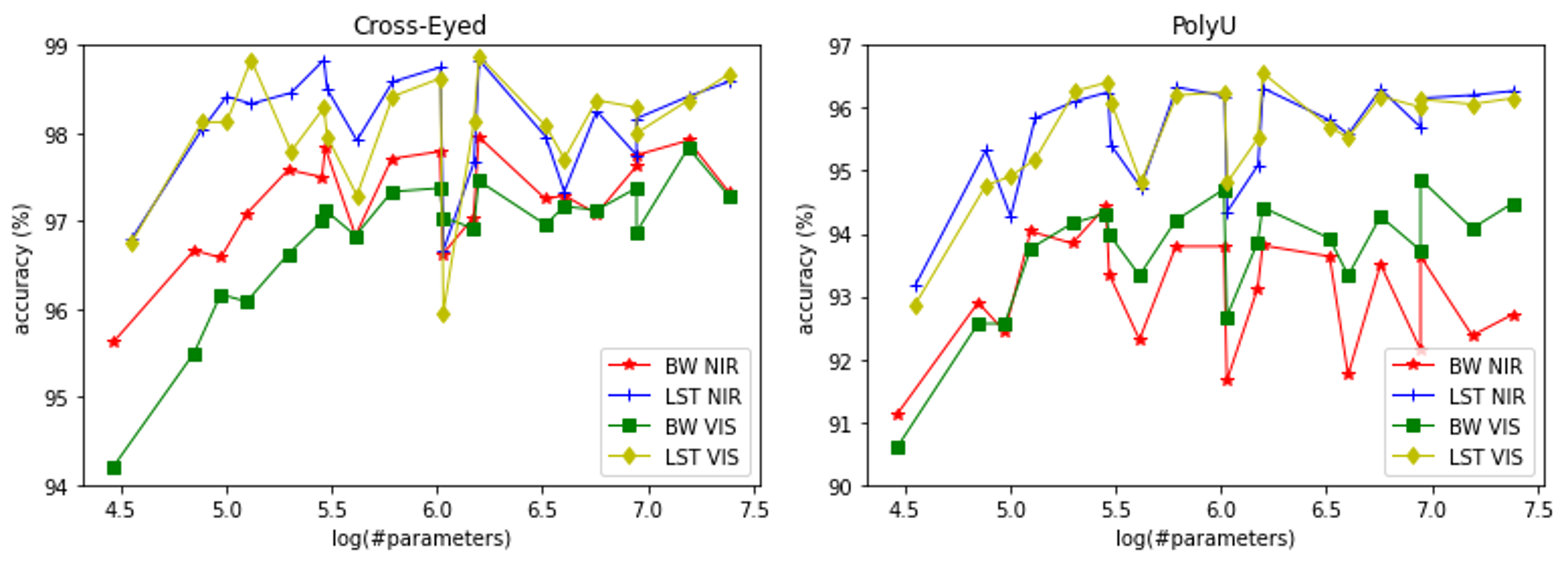}
\caption{Average test accuracy for different network compression levels. The number of parameters depends on the width of the network, controlled by coefficient $\alpha$, and the depth of the network, as explained in Section \ref{compression}. The complex data representation used at the input is $\Re(I_{20})$,$\Im(I_{20})$, $I_{11}$. 
%parameter effect (point is 256 full size network and alpha=0.2.
\label{parametereffect}}
\end{figure}

Since ready-to-use features are provided up-front, there is potential for network compression and speed improvement. Thus, we also experimented with different network depths, widths, and input resolutions. A coefficient $\alpha$ was used to reduce the number of filters per layer (width) of ResNet50 with values ($0.2$, $0.4$, $0.6$, $0.8$, and $1.0$), being $1.0$ the standard network size. In addition, the input image size was reduced by one-fourth each time while simultaneously changing the network depth to compensate, going from an input size of 256x256 pixels to 32x32 pixels in its shallowest version. We maintained the network's abstraction level and output resolution by decreasing the depth of the network accordingly.

%For each of the image resolutions, the network depth was decreased accordingly by cutting off the network’s end just before the next reduction in feature space resolution. This way, we maintained the network’s abstraction level and output resolution.

Figure \ref{parametereffect} shows the effect that the amount of parameters has on performance. The amount of parameters depends on the network's width and depth of the network. The first thing to notice is that all curves exhibit very similar behavior regardless of the database, spectrum, and input type. The slight offset in terms of the number of parameters in comparison to grayscale images is due to the increase in the number of channels at the input. This offset ($6,272$ parameters) is minor compared to the number of parameters of the original ResNet50 (25.5 million). 
It can also be seen that CST outperforms the grayscale images in all cases, except for one point in the Cross-Eyed dataset and VIS spectrum at around 1 million parameters. At this point, the network depth is at its maximum, but the width is at its minimum ($\alpha=0.2$). Such extreme imbalance is the explanation for the sudden drop in performance observed in the graphs for both datasets. Another observation from the graphs is that the network can go through compression to a large extent without loss in performance, with our solution systematically providing better accuracy for any compression level. This suggests that our method could support compression, producing smaller networks while still benefiting from an increase in performance.

\subsection{Consistency across other CNN architectures}

\begin{table*}[h!]
\centering
\resizebox{0.7\textwidth}{!}{%
\begin{tabular}{c|cccc|cccc|}
\cline{2-9}
 &
  \multicolumn{4}{c|}{\textbf{Cross-Eyed}} &
  \multicolumn{4}{c|}{\textbf{PolyU}} \\ \cline{2-9} 
 &
  \multicolumn{2}{c|}{\textbf{NIR}} &
  \multicolumn{2}{c|}{\textbf{VIS}} &
  \multicolumn{2}{c|}{\textbf{NIR}} &
  \multicolumn{2}{c|}{\textbf{VIS}} \\ \hline
\multicolumn{1}{|c|}{\textbf{Network}} &
  \multicolumn{1}{c|}{\textbf{BW}} &
  \multicolumn{1}{c|}{\textbf{CST}} &
  \multicolumn{1}{c|}{\textbf{BW}} &
  \textbf{CST} &
  \multicolumn{1}{c|}{\textbf{BW}} &
  \multicolumn{1}{c|}{\textbf{CST}} &
  \multicolumn{1}{c|}{\textbf{BW}} &
  \textbf{CST} \\ \hline
\multicolumn{1}{|c|}{\textbf{ResNet50}} &
  \multicolumn{1}{c|}{97.8} &
  \multicolumn{1}{c|}{\textbf{98.5}} &
  \multicolumn{1}{c|}{97.7} &
  \textbf{98.6} &
  \multicolumn{1}{c|}{93.3} &
  \multicolumn{1}{c|}{\textbf{96.3}} &
  \multicolumn{1}{c|}{94.5} &
  \textbf{96.4} \\ \hline
\multicolumn{1}{|c|}{\textbf{DenseNet121}} &
  \multicolumn{1}{c|}{98.3} &
  \multicolumn{1}{c|}{\textbf{99.1}} &
  \multicolumn{1}{c|}{97.8} &
  \textbf{99.3} &
  \multicolumn{1}{c|}{95.8} &
  \multicolumn{1}{c|}{\cellcolor[HTML]{C0C0C0}\textbf{98.1}} &
  \multicolumn{1}{c|}{97.2} &
  \textbf{98.0} \\ \hline
\multicolumn{1}{|c|}{\textbf{Xception}} &
  \multicolumn{1}{c|}{98.5} &
  \multicolumn{1}{c|}{\cellcolor[HTML]{C0C0C0}\textbf{99.3}} &
  \multicolumn{1}{c|}{98.9} &
  \cellcolor[HTML]{C0C0C0}\textbf{99.6} &
  \multicolumn{1}{c|}{96.2} &
  \multicolumn{1}{c|}{\textbf{98.0}} &
  \multicolumn{1}{c|}{97.4} &
  \cellcolor[HTML]{C0C0C0}\textbf{98.1} \\ \hline
\multicolumn{1}{|c|}{\textbf{InceptionV3}} &
  \multicolumn{1}{c|}{97.8} &
  \multicolumn{1}{c|}{\textbf{98.9}} &
  \multicolumn{1}{c|}{97.9} &
  \textbf{98.3} &
  \multicolumn{1}{c|}{93.8} &
  \multicolumn{1}{c|}{\textbf{96.5}} &
  \multicolumn{1}{c|}{94.9} &
  \textbf{95.9} \\ \hline
\multicolumn{1}{|c|}{\textbf{MobileNetV2}} &
  \multicolumn{1}{c|}{98.6} &
  \multicolumn{1}{c|}{\textbf{99.0}} &
  \multicolumn{1}{c|}{98.4} &
  \textbf{99.2} &
  \multicolumn{1}{c|}{94.5} &
  \multicolumn{1}{c|}{\textbf{97.3}} &
  \multicolumn{1}{c|}{96.1} &
  \textbf{97.4} \\ \hline
\end{tabular}
}
\caption{\label{tab:best_network} Average test accuracy for each network. The complex data representation used at the input is $\Re(I_{20})$,$\Im(I_{20})$, $I_{11}$. The bold numbers indicate the best case among BW (grayscale images) and CST (our approach). The gray cells indicate the best result per column. 
}
\end{table*}

% We have also applied the proposed method to other popular CNN architectures to corroborate its benefits. Table \ref{tab:best_network} shows the results of such experimentation. It can be seen that for all networks and databases, our complex structure tensor version outperforms the baseline grayscale version. For the Cross-Eyed database, our VGG16 version improves accuracy by $3,6\%$ and $2.3\%$ for the NIR and VIS case, respectively. For PolyU, our ResNet50 network improves the baseline results by $3,3\%$ for the NIR case and $1,9\%$ for VIS. It is also worth noting the improvements of InceptionV3 and MobileNetV2 with PolyU-NIR ($2.7\%$ and $2.8\%$, respectively). In absolute terms, the best-performing network for both the baseline and our method is Xception, reaching up to $99,6\%$ accuracy for the VIS Cross-Eyed data and $98\%-99\%$ accuracy in all other cases. In summary, the observed accuracy improvements with our solution demonstrate that it also generalizes to other CNN architectures.

We have also applied the proposed method to other popular CNN architectures to corroborate its benefits. Table \ref{tab:best_network} shows the results of such experimentation. It can be seen that for all networks and databases, our complex structure tensor version outperforms the baseline grayscale version. For the Cross-Eyed database. For PolyU, our ResNet50 network improves the baseline results by $3.3\%$ for the NIR case and $1.9\%$ for VIS. It is also worth noting the improvements of InceptionV3 and MobileNetV2 with PolyU-NIR ($2.7\%$ and $2.8\%$, respectively). In absolute terms, the best-performing network for both the baseline and our method is Xception, reaching up to $99,6\%$ accuracy for the VIS Cross-Eyed data and $98\%-99\%$ accuracy in all other cases. In summary, the observed accuracy improvements with our solution demonstrate that it also generalizes to other CNN architectures.

\subsection{Performance comparison with previous studies}
%\subsection{Comparison with previous studies}

\begin{figure}[!h]
\centering
\includegraphics[width=0.95\columnwidth]{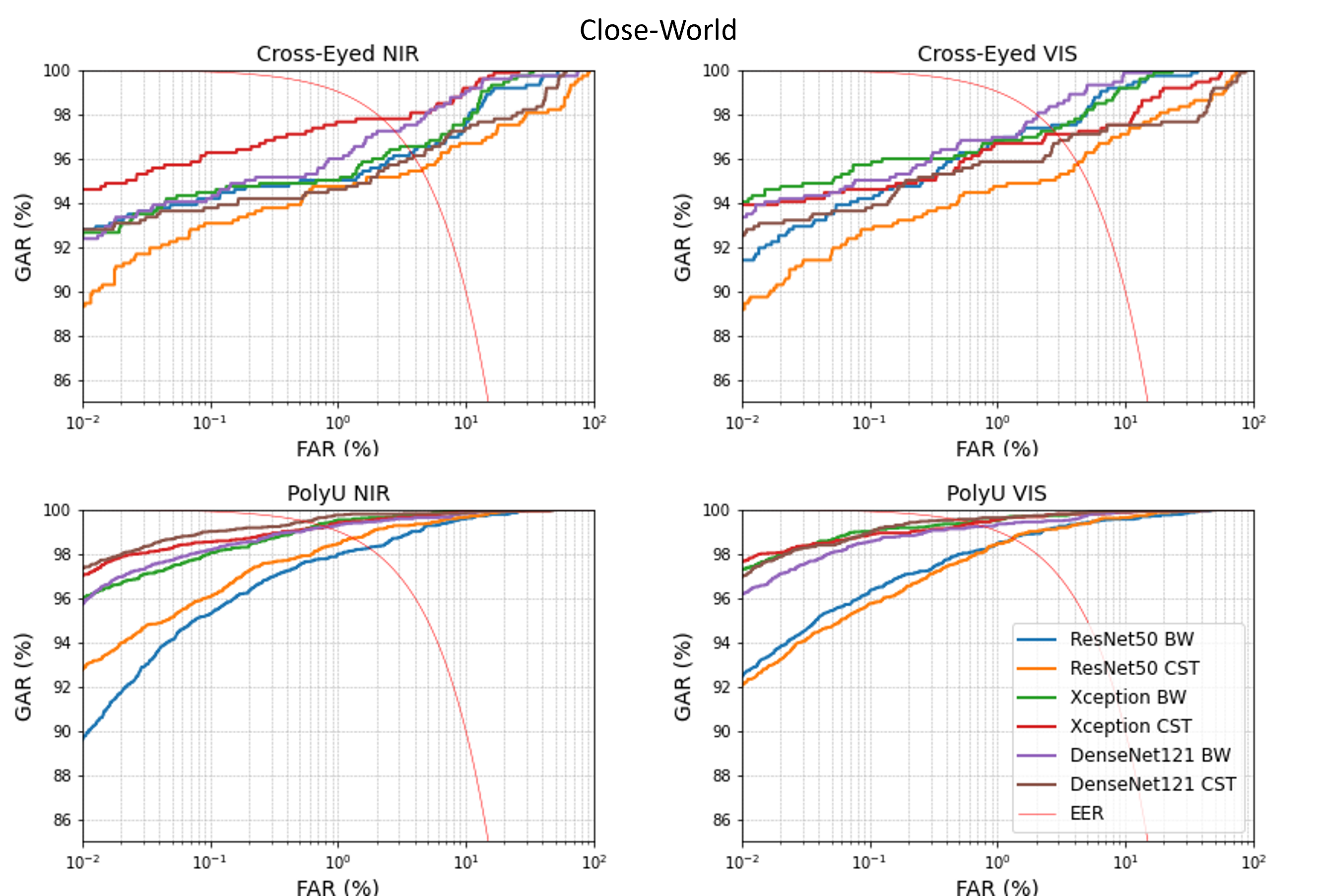}
\caption{ROC curves for the Close-World setting. The EER values and comparison with the SOA results are given in Table \ref{tab:SOA}
\label{ROC_CW}}
\end{figure}

\begin{figure}[!h]
\centering
\includegraphics[width=0.95\columnwidth]{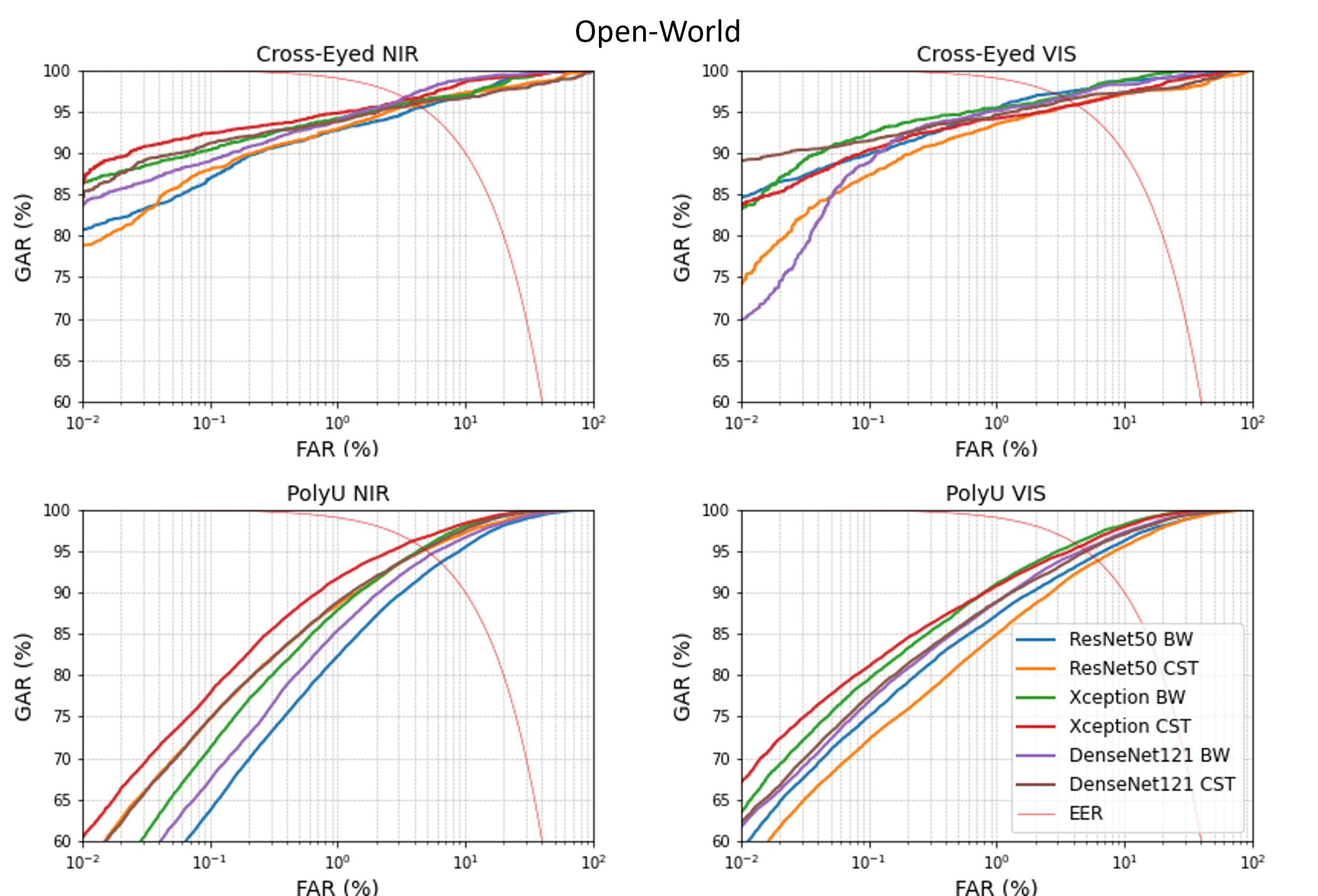}
\caption{ROC curves for the Open-World setting. The EER values and comparison with the SOA results are given in Table \ref{tab:SOA}
\label{ROC_OW}}
\end{figure}

Table \ref{tab:SOA} summarizes and compares our results with previous works using the same databases \cite{hernandez2023one}\cite{depresion}. The whole performance for each system can be seen in the ROC curves presented in Figures \ref{ROC_CW} and \ref{ROC_OW} for the Close-World and Open-World cases, respectively. This sub-section follows the same protocol as those previous works to enable the comparison as explained in Section \ref{metricsprotocol}. 
%In particular, we apply a biometric verification scenario under the CW and OW protocols of Table \ref{tabla-gen-imp}, as described in Section \ref{metricsprotocol}. 
The CST method in these experiments used the input combination of BW, $\Re(I_{20})$,$\Im(I_{20})$, $I_{11}$. Including the grayscale image in these experiments proposes a more general approach of using CST with Deep Learning. 

\begin{table}[h!]
\centering
\resizebox{0.85\columnwidth}{!}{%
\begin{tabular}{cc|cccc|cccc|}
\cline{3-10}
  &
    &
    
   \multicolumn{2}{c|}{\textbf{Cross-Eyed (CW)}} &
   \multicolumn{2}{c|}{\textbf{PolyU (CW)}} &
   \multicolumn{2}{c|}{\textbf{Cross-Eyed (OW)}} &
   \multicolumn{2}{c|}{\textbf{PolyU (OW)}} \\ \cline{3-10}
  &
    &
   \multicolumn{1}{c|}{\textbf{NIR}} &
   \multicolumn{1}{c|}{\textbf{VIS}} &
   \multicolumn{1}{c|}{\textbf{NIR}} &
   \textbf{VIS} &
   \multicolumn{1}{c|}{\textbf{NIR}} &
   \multicolumn{1}{c|}{\textbf{VIS}} &
   \multicolumn{1}{c|}{\textbf{NIR}} &
   \textbf{VIS} \\ \hline
\multicolumn{1}{|c|}{\multirow{4}{*}{\textbf{ResNet50}}} &
   \textbf{BW} &
   \multicolumn{1}{c|}{3.7} &
   \multicolumn{1}{c|}{2.6} &
   \multicolumn{1}{c|}{1.8} &
   1.3 &
   \multicolumn{1}{c|}{4.5} &
   \multicolumn{1}{c|}{\textbf{2.9}} &
   \multicolumn{1}{c|}{6.4} &
   5.7 \\ \cline{2-10}
\multicolumn{1}{|c|}{} &
   \textbf{CST} &
   \multicolumn{1}{c|}{4.6} &
   \multicolumn{1}{c|}{4.3} &
   \multicolumn{1}{c|}{\cellcolor[HTML]{C0C0C0}1.3} &
   \cellcolor[HTML]{C0C0C0}1.3 &
   \multicolumn{1}{c|}{\cellcolor[HTML]{C0C0C0}4.3} &
   \multicolumn{1}{c|}{4.2} &
   \multicolumn{1}{c|}{\cellcolor[HTML]{C0C0C0}4.9} &
   6.2 \\ \cline{2-10}
\multicolumn{1}{|c|}{} &
   \cite{depresion} &
   \multicolumn{1}{c|}{\textbf{1.8}} &
   \multicolumn{1}{c|}{1.7} &
   \multicolumn{1}{c|}{0.68} &
   0.61 &
   \multicolumn{1}{c|}{3.5} &
   \multicolumn{1}{c|}{3.4} &
   \multicolumn{1}{c|}{4.00} &
   \textbf{3.94} \\ \cline{2-10}
\multicolumn{1}{|c|}{} &
   \cite{hernandez2023one} &
   \multicolumn{1}{c|}{-} &
   \multicolumn{1}{c|}{\textbf{1.3}} &
   \multicolumn{1}{c|}{-} &
   3.4 &
   \multicolumn{1}{c|}{-} &
   \multicolumn{1}{c|}{\textbf{1.2}} &
   \multicolumn{1}{c|}{-} &
   7.76 \\ \hline
\multicolumn{1}{|c|}{\multirow{2}{*}{\textbf{Xception}}} &
   \textbf{BW} &
   \multicolumn{1}{c|}{3.5} &
   \multicolumn{1}{c|}{2.8} &
   \multicolumn{1}{c|}{0.7} &
   0.6 &
   \multicolumn{1}{c|}{3.8} &
   \multicolumn{1}{c|}{3.2} &
   \multicolumn{1}{c|}{4.7} &
   \textbf{4.1} \\ \cline{2-10}
\multicolumn{1}{|c|}{} &
   \textbf{CST} &
   \multicolumn{1}{c|}{\cellcolor[HTML]{C0C0C0}\textbf{2.2}} &
   \multicolumn{1}{c|}{2.9} &
   \multicolumn{1}{c|}{0.8} &
   \cellcolor[HTML]{C0C0C0}0.6 &
   \multicolumn{1}{c|}{\cellcolor[HTML]{C0C0C0}3.5} &
   \multicolumn{1}{c|}{4.2} &
   \multicolumn{1}{c|}{\cellcolor[HTML]{C0C0C0}\textbf{3.8}} &
   4.5 \\ \hline
\multicolumn{1}{|c|}{\multirow{3}{*}{\textbf{DenseNet121}}} &
   \textbf{BW} &
   \multicolumn{1}{c|}{2.8} &
   \multicolumn{1}{c|}{\textbf{2.1}} &
   \multicolumn{1}{c|}{0.8} &
   0.8 &
   \multicolumn{1}{c|}{\textbf{3.3}} &
   \multicolumn{1}{c|}{3.4} &
   \multicolumn{1}{c|}{5.4} &
   4.9 \\ \cline{2-10}
\multicolumn{1}{|c|}{} &
   \textbf{CST} &
   \multicolumn{1}{c|}{3.9} &
   \multicolumn{1}{c|}{3.2} &
   \multicolumn{1}{c|}{\cellcolor[HTML]{C0C0C0}\textbf{0.5}} &
   \cellcolor[HTML]{C0C0C0}\textbf{0.5} &
   \multicolumn{1}{c|}{4.1} &
   \multicolumn{1}{c|}{3.6} &
   \multicolumn{1}{c|}{\cellcolor[HTML]{C0C0C0}4.8} &
   5.0 \\ \cline{2-10}
\multicolumn{1}{|c|}{} &
   \cite{hernandez2023one} &
   \multicolumn{1}{c|}{-} &
   \multicolumn{1}{c|}{1.6} &
   \multicolumn{1}{c|}{-} &
   2.49 &
   \multicolumn{1}{c|}{-} &
   \multicolumn{1}{c|}{1.7} &
   \multicolumn{1}{c|}{-} &
   5.82 \\ \hline
\end{tabular}
}
\caption{\label{tab:SOA}
CNN networks comparison with absence and presence of CST input, as well as SOA EERs for each database and spectrum. Bold cells show the best-overall and our best result for each particular setting (column). Shaded cells emphasize improvement with the presence of the CST over using only grayscale images.
   %$^{*}$ values are taken from \cite{hernandez2023one}
}
\end{table}

In this case, results are more mixed than in the previous subsections. In particular, the CST approach does not outperform the use of original grayscale images in all cases. This contrast can be partially attributed to the use of 5-fold cross-validation in the previous sections, which shows a more reliable indication of the performance of the system by testing over the whole dataset. 
%In this case, we only calculate the EER on a single test partition (given by the literature), which can give a slightly skewed view of the method's overall performance. 
This effect mostly happens with the Cross-Eyed database. We attribute this to the more limited training data available in Cross-Eyed since the networks are trained from scratch, and our CST approach involves a higher amount of input channels. 
Conversely, with PolyU, our approach works better or the same as the original grayscale images in most cases.
%Conversely, with PolyU, the amount of training data is higher, so our approach works better than the original grayscale images or at least almost the same in most cases. Therefore, with sufficient training data, our method can show superior performance in a verification setting.

From the ROC curves in Figures \ref{ROC_CW} and \ref{ROC_OW}, one can see that in some cases, despite a higher EER in Table \ref{tab:SOA}, the performance favors ConvNet models using CST information when ROC curves enter the low FAR area. Such cases are the Xception network on the PolyU dataset for both the NIR case and Close-World Protocol, as well as for the VIS spectrum in the Open-World Protocol. 

For the Open-World case, our CST model performs better at a lower FAR for every case, with the exception of PolyU in the VIS spectrum, where the difference between models is negligible. For DenseNet121, the performance difference in Cross-Eyed and NIR for the Close-World setting becomes smaller at lower FAR. Despite the variability in the results due to single-partition testing, ROC curve results confirm predictions of the identification sections. Texture orientation information brings measurable performance benefits in recognition compared to using only grayscale information when networks were not pre-trained at all.

% update this
The results of Table \ref{tab:SOA} indicate the EER of CST and the SOA results from previous studies \cite{hernandez2023one}\cite{depresion} in the same protocol. Despite CST having no access to pre-training, it is able to deliver SOA performance with Xception and DenseNet121 networks in PolyU in CW (both spectra) and in the difficult OW with PolyU in NIR. In the case of ResNet50, which delivers the best performance of state-of-the-art in the protocol, CST was able to come closer than $1$ percentage point to the state-of-the-art in CW, PolyU NIR CW, and in the difficult OW PolyU NIR cases. In other cases, CST was lagging behind by not more than $2$-$3$ percentage points.
The method of \cite{depresion} used a ResNet50 network pre-trained for face recognition on the very large VGGFace2 and MS1M databases. 
This leads us to hypothesize that our strategy of using mini complex ConvNets to complement grayscale with texture orientation at the input of CNNs could also benefit from pre-training on other modalities, such as face, where much larger databases exist. 
On the other hand, the method of \cite{hernandez2023one} underwent an exhaustive search to find layers of the networks that could provide enhanced verification performance. %Nonetheless, our results evidence that when huge ConvNets starve for more data in the training, network coefficients produced by texture orientation inputs are consistently closer to the optimum (beyond the reach due to scarce data) than not offering them to the network upfront.

\section{Conclusions}\label{Conclusions}

This study examined the benefits of the Complex Structure Tensor theory to improve CNNs by providing measurements representing the presence of unique texture orientation and its angle at the input. We investigated the effect of changing the input image of the network from the original grayscale to a preprocessed image where the joint presence of up to $n$ orientations in the pixels' neighborhood is highlighted. This is achieved via second-order complex moments of local power spectra, which produce a complex-value representation of the image that encodes both the magnitude and angle of the dominant texture (group) orientations. We used periocular recognition as a case study and six well-known network architectures. We employed two common datasets used in periocular research, Cross-Eyed and PolyU, which contain images both in near-infrared (NIR) and visible (VIS) spectra. This allowed us to explore the applicability of our findings to different spectra.

Typically a Gabor filter bank representing the local spectrum is obtained by using 30-100 (=dimension) different filters. In our method, we extract single frequency band texture components in the image, which are then averaged over a neighborhood (configurable via two parameters, $\sigma_1$ and $\sigma_2$, respectively). This approach compresses the feature space dimension of the image that is provided to the network while maintaining translation invariant information of the local power-spectra, improving accuracy and convergence.
%However, our results show that sufficiently meaningful textural information is retained despite such a reduction, improving identification accuracy and network convergence compared to using original grayscale images. 
Improvement is observed regardless of the database, spectrum, or network architecture. This suggests that standard CNNs are not able to effectively reach these features from the gray-scale images, especially when training data is limited.

%The mentioned reduction of the feature space led us to explore potential reductions in network depth (layers) and width (filters per layer). 
We show that, with our method, thanks to the reduction of the feature space, network compression is possible while maintaining superior performance. In terms of comparison with prior studies, our method is shown to improve the verification performance when sufficient training data is available.

With orientation features at the input, we also gained explainability as to CNNs' decision basis. Networks can then focus on what has proven helpful for mammalian vision upfront. Even if CNNs have been shown to extract select texture orientations in first layers, \cite{gabornet}\cite{krizhevsky2012imagenet}, our results indicate that providing them upfront reduced the burden of learning and extracting them with higher mathematical systemacy, at no cost (downsizable network). The systemacy included \cite{zhang2002brief} mathematical completeness, interpolative direction, independence of filter bank size for direction accuracy, certainty measure, and translation invariance within textures.
%In a similar vein to ours, recent works \cite{luan2018gabor}\cite{Per20} replaced the first layers of CNNs with learnable Gabor filters, observing similar benefits in terms of accuracy and convergence. More strikingly, the first layers of popular CNNs \cite{krizhevsky2012imagenet} converge into Gabor-like filters. Thus, incorporating biologically inspired models into deep learning solutions can give significant benefits, but this direction is highly unexplored \cite{sun2017demographic}. From our experiments, such low-level biologically-inspired information seems unreachable or difficult to access by the networks if not made explicit. 

% modify this last paragraph
% To fully leverage the presented method's benefits, we aim to go further by making the hyperparameters of the employed filters learnable. Symmetry filters, built from derivatives of Gaussians and specialized in texture orientations, are defined by a few parameters, which have the potential to speed up training. In addition, choosing filter family parameters (and not filter values) will contribute to layer interpretability since the families employed have well-known interpretations and goodness of the fitted model. Furthermore, we can combine the filter families to extract specific symmetry patterns as in \cite{symmetryfingerprint} in combination with texture to improve detection and segmentations of specific patterns in images such as the iris region.

To fully leverage the presented method's benefits, we aim to go further by making the hyperparameters of the employed filters learnable. Our use of local texture features and the increase in convergence, network compression, and the separability of the filters used, make our model an interesting case for newer architectures like hybrid vision transformers, where recent works focus on models' accuracy-latency trade-off \cite{vasufastvit2023}. Future work also includes applications in CV where texture information and/or certainty play important roles, such as image segmentation \cite{semantic}, or image generation \cite{yang2023diffusion}.

\section{Acknowledgements}

This work was supported by Vetenskapsrådet [grant numbers 2016-03497 and 2021-05110] and VINNOVA [grant number 2022-00919].

%% The Appendices part is started with the command \appendix;
%% appendix sections are then done as normal sections
% \appendix

% \section{Sample Appendix Section}
% \label{sec:sample:appendix}
% Lorem ipsum dolor sit amet, consectetur adipiscing elit, sed do eiusmod tempor section \ref{sec:sample1} incididunt ut labore et dolore magna aliqua. Ut enim ad minim veniam, quis nostrud exercitation ullamco laboris nisi ut aliquip ex ea commodo consequat. Duis aute irure dolor in reprehenderit in voluptate velit esse cillum dolore eu fugiat nulla pariatur. Excepteur sint occaecat cupidatat non proident, sunt in culpa qui officia deserunt mollit anim id est laborum.

%% If you have bibdatabase file and want bibtex to generate the
%% bibitems, please use
%%
 \bibliographystyle{elsarticle-num} 
 \bibliography{cas-refs}

\begin{thebibliography}{10}
\expandafter\ifx\csname url\endcsname\relax
  \def\url#1{\texttt{#1}}\fi
\expandafter\ifx\csname urlprefix\endcsname\relax\def\urlprefix{URL }\fi
\expandafter\ifx\csname href\endcsname\relax
  \def\href#1#2{#2} \def\path#1{#1}\fi

\bibitem{tuceryan1993texture}
M.~Tuceryan, A.~K. Jain, Texture analysis, Handbook of pattern recognition and computer vision (1993) 235--276.

\bibitem{zhang2002brief}
J.~Zhang, T.~Tan, Brief review of invariant texture analysis methods, Pattern recognition 35~(3) (2002) 735--747.

\bibitem{julesztexton}
B.~Julesz, J.~Bergen, Textons, the fundemantal elements in preattentive vision and perception of textures, The bell system technical journal 62~(6) (1983) 1619--1645, %place: psychophysics.

\bibitem{HOG}
N.~Dalal, B.~Triggs, Histograms of oriented gradients for human detection, in: 2005 IEEE computer society conference on computer vision and pattern recognition (CVPR'05), Vol.~1, Ieee, 2005, pp. 886--893.

\bibitem{LBP}
T.~Ojala, M.~Pietikainen, D.~Harwood, Performance evaluation of texture measures with classification based on kullback discrimination of distributions, in: Proceedings of 12th international conference on pattern recognition, Vol.~1, IEEE, 1994, pp. 582--585.

\bibitem{geirhos2018imagenettrained}
R.~Geirhos, P.~Rubisch, C.~Michaelis, M.~Bethge, F.~A. Wichmann, W.~Brendel, Imagenet-trained {CNN}s are biased towards texture; increasing shape bias improves accuracy and robustness., in: International Conference on Learning Representations, 2019.

\bibitem{krizhevsky2012imagenet}
A.~Krizhevsky, I.~Sutskever, G.~E. Hinton, Imagenet classification with deep convolutional neural networks, Advances in neural information processing systems 25 (2012).

\bibitem{hubel59}
D.~Hubel, T.~Wiesel, Receptive fields of single neurons in the cat's striate cortex, J. physiol. (London) 148 (1959) 574--591.

\bibitem{gabor}
G.~H. Granlund, In search of a general picture processing operator, Computer Graphics and Image Processing 8~(2) (1978) 155--173.

\bibitem{kulikowski81}
J.~Kulikowski, P.~Bishop, Fourier analysis and spatial representation in the visual cortex, Experientia 37 (1981) 160--162.

\bibitem{jain2001fingerprint}
A.~Jain, A.~Ross, S.~Prabhakar, Fingerprint matching using minutiae and texture features, in: Proceedings 2001 International Conference on Image Processing (Cat. No. 01CH37205), Vol.~3, IEEE, 2001, pp. 282--285.

\bibitem{daugman2009iris}
J.~Daugman, How iris recognition works, in: The essential guide to image processing, Elsevier, 2009, pp. 715--739.

\bibitem{park2009periocular}
U.~Park, A.~Ross, A.~K. Jain, Periocular biometrics in the visible spectrum: A feasibility study, in: 2009 IEEE 3rd international conference on biometrics: theory, applications, and systems, IEEE, 2009, pp. 1--6.

\bibitem{alonso2016survey}
F.~Alonso-Fernandez, J.~Bigun, A survey on periocular biometrics research, Pattern Recognition Letters 82 (2016) 92--105.

\bibitem{periocularlbp}
L.~C.~O. Tiong, Y.~Lee, A.~B.~J. Teoh, Periocular recognition in the wild: Implementation of rgb-oclbcp dual-stream cnn, Applied Sciences 9~(13) (2019) 2709.

\bibitem{hernandez2023one}
K.~Hernandez-Diaz, F.~Alonso-Fernandez, J.~Bigun, One-shot learning for periocular recognition: Exploring the effect of domain adaptation and data bias on deep representations, IEEE Access (2023).

\bibitem{bigun06vd}
J.~Bigun, Vision with Direction, Springer, Heidelberg, 2006.
\newblock \href {https://doi.org/10.1007/b138918} {\path{doi:10.1007/b138918}}.

\bibitem{bigun87london}
J.~Bigun, G.~Granlund, Optimal orientation detection of linear symmetry, in: {First ICCV}, {London, June 8--11}, IEEE Computer Society, 1987, pp. 433--438.

\bibitem{weickert1999coherence}
J.~Weickert, Coherence-enhancing diffusion filtering, International journal of computer vision 31 (1999) 111--127.

\bibitem{szczepankiewicz2016link}
F.~Szczepankiewicz, D.~van Westen, E.~Englund, C.-F. Westin, F.~St{\aa}hlberg, J.~L{\"a}tt, P.~C. Sundgren, M.~Nilsson, The link between diffusion mri and tumor heterogeneity: Mapping cell eccentricity and density by diffusional variance decomposition (divide), Neuroimage 142 (2016) 522--532.

\bibitem{lindeberg2013scale}
T.~Lindeberg, Scale-space theory in computer vision, Vol. 256, Springer Science \& Business Media, 2013.

\bibitem{mikolajczyk}
K.~Mikolajczyk, C.~Schmid, Scale \& affine invariant interest point detectors, International journal of computer vision 60~(1) (2004) 63--86.

\bibitem{bruhn2005lucas}
A.~Bruhn, J.~Weickert, C.~Schn{\"o}rr, Lucas/kanade meets horn/schunck: Combining local and global optic flow methods, International Journal of Computer Vision 61~(3) (2005) 211--231.

\bibitem{si2017dense}
X.~Si, J.~Feng, B.~Yuan, J.~Zhou, Dense registration of fingerprints, Pattern Recognition 63 (2017) 87--101.

\bibitem{symmetryfingerprint}
A.~Mikaelyan, F.~Alonso-Fernandez, J.~Bigun, Keypoint description by symmetry assessment -- applications in biometrics (2023).
\newblock \href {http://arxiv.org/abs/2311.01651} {\path{arXiv:2311.01651}}.

\bibitem{gabornet}
A.~Alekseev, A.~Bobe, Gabornet: Gabor filters with learnable parameters in deep convolutional neural network, in: 2019 International Conference on Engineering and Telecommunication (EnT), IEEE, 2019, pp. 1--4.

\bibitem{LBPCNNfeatures}
R.~P. Singh, R.~Dash, R.~K. Mohapatra, Lbp and cnn feature fusion for face anti-spoofing, Pattern Analysis and Applications 26~(2) (2023) 773--782.

\bibitem{LBPNet}
F.~Juefei-Xu, V.~Naresh~Boddeti, M.~Savvides, Local binary convolutional neural networks, in: Proceedings of the IEEE conference on computer vision and pattern recognition, 2017, pp. 19--28.

\bibitem{HOGCNNtracking}
L.~Kalake, Y.~Dong, W.~Wan, L.~Hou, Enhancing detection quality rate with a combined hog and cnn for real-time multiple object tracking across non-overlapping multiple cameras, Sensors 22~(6) (2022) 2123.

\bibitem{hogcnnperiocular}
P.~Kumari, K.~Seeja, A novel periocular biometrics solution for authentication during covid-19 pandemic situation, Journal of Ambient Intelligence and Humanized Computing 12 (2021) 10321--10337.

\bibitem{hogcnnexpression}
X.~Pan, Fusing hog and convolutional neural network spatial--temporal features for video-based facial expression recognition, IET Image Processing 14~(1) (2020) 176--182.

\bibitem{gaborensemble}
J.~Y. Choi, B.~Lee, Ensemble of deep convolutional neural networks with gabor face representations for face recognition, IEEE Transactions on Image Processing 29 (2019) 3270--3281.

\bibitem{gaboremotions}
M.~M.~T. Zadeh, M.~Imani, B.~Majidi, Fast facial emotion recognition using convolutional neural networks and gabor filters, in: 2019 5th Conference on Knowledge Based Engineering and Innovation (KBEI), IEEE, 2019, pp. 577--581.

\bibitem{gaborcapsule}
S.~Hosseini, N.~I. Cho, Gf-capsnet: Using gabor jet and capsule networks for facial age, gender, and expression recognition, in: 2019 14th IEEE International Conference on Automatic Face \& Gesture Recognition (FG 2019), IEEE, 2019, pp. 1--8.

\bibitem{resnet}
K.~He, X.~Zhang, S.~Ren, J.~Sun, Deep residual learning for image recognition, in: Proceedings of the IEEE conference on computer vision and pattern recognition, 2016, pp. 770--778.

\bibitem{densenet}
G.~Huang, Z.~Liu, L.~Van Der~Maaten, K.~Q. Weinberger, Densely connected convolutional networks, in: Proceedings of the IEEE conference on computer vision and pattern recognition, 2017, pp. 4700--4708.

\bibitem{xception}
F.~Chollet, Xception: Deep learning with depthwise separable convolutions, in: Proceedings of the IEEE conference on computer vision and pattern recognition, 2017, pp. 1251--1258.

\bibitem{inception}
C.~Szegedy, V.~Vanhoucke, S.~Ioffe, J.~Shlens, Z.~Wojna, Rethinking the inception architecture for computer vision, in: Proceedings of the IEEE conference on computer vision and pattern recognition, 2016, pp. 2818--2826.

\bibitem{mobilenet}
M.~Sandler, A.~Howard, M.~Zhu, A.~Zhmoginov, L.-C. Chen, Mobilenetv2: Inverted residuals and linear bottlenecks, in: Proceedings of the IEEE conference on computer vision and pattern recognition, 2018, pp. 4510--4520.

\bibitem{depresion}
L.~A. Zanlorensi, D.~R. Lucio, A.~d.~S. Britto~Junior, H.~Proen{\c{c}}a, D.~Menotti, Deep representations for cross-spectral ocular biometrics, IET Biometrics 9~(2) (2020) 68--77.

\bibitem{x-eyed2016}
A.~Sequeira, L.~Chen, P.~Wild, J.~Ferryman, F.~Alonso-Fernandez, K.~B. Raja, R.~Raghavendra, C.~Busch, J.~Bigun, Cross-eyed-cross-spectral iris/periocular recognition database and competition, in: 2016 International Conference of the Biometrics Special Interest Group (BIOSIG), IEEE, 2016, pp. 1--5.

\bibitem{polyu}
P.~R. Nalla, A.~Kumar, Toward more accurate iris recognition using cross-spectral matching, IEEE transactions on Image processing 26~(1) (2016) 208--221.

\bibitem{vasufastvit2023}
P.~K.~A. Vasu, J.~Gabriel, J.~Zhu, O.~Tuzel, A.~Ranjan, Fastvit: A fast hybrid vision transformer using structural reparameterization, in: Proceedings of the IEEE/CVF International Conference on Computer Vision, 2023.

\bibitem{semantic}
Y.~Mo, Y.~Wu, X.~Yang, F.~Liu, Y.~Liao, Review the state-of-the-art technologies of semantic segmentation based on deep learning, Neurocomputing 493 (2022) 626--646.

\bibitem{yang2023diffusion}
L.~Yang, Z.~Zhang, Y.~Song, S.~Hong, R.~Xu, Y.~Zhao, W.~Zhang, B.~Cui, M.-H. Yang, Diffusion models: A comprehensive survey of methods and applications, ACM Computing Surveys 56~(4) (2023) 1--39.

\end{thebibliography}

%% else use the following coding to input the bibitems directly in the
%% TeX file.

% \begin{thebibliography}{00}

% %% \bibitem{label}
% %% Text of bibliographic item

% \bibitem{}

% \end{thebibliography}
\end{document}